\documentclass[letterpaper, 10 pt, journal, twoside]{ieeeconf}  % Comment this line out if you need a4paper

\IEEEoverridecommandlockouts                              % This command is only needed if 
\let\labelindent\relax
\usepackage{multirow}
\usepackage{amsmath} % assumes amsmath package installed
\usepackage{amssymb}  % assumes amsmath package installed
\usepackage{amsfonts}
\usepackage{amsthm}
\usepackage{graphicx}
\usepackage{hyperref}
\usepackage{booktabs}
\usepackage{nicefrac}
\usepackage{enumitem}
\usepackage{algorithm, algcompatible}
\usepackage{pifont}
\usepackage{xcolor} % rebuttal

\newcommand{\tabref}[1]{Tab.~\ref{#1}}
\newcommand{\equref}[1]{Eq.~\ref{#1}}
\newcommand{\figref}[1]{Fig.~\ref{#1}}
\newcommand{\secref}[1]{Sec.~\ref{#1}}

\algnewcommand\INPUT{\item[\textbf{Input:}]}%
\algnewcommand\OUTPUT{\item[\textbf{Output:}]}%

\title{\LARGE \bf
Scale-invariant and View-relational Representation Learning for Full Surround Monocular Depth
}

\author{Kyumin Hwang$^{1,*}$, Wonhyeok Choi$^{1,*}$, Kiljoon Han$^{1}$, Wonjoon Choi$^{1}$, Minwoo Choi$^{1}$,\\Yongcheon Na$^{2}$, Minwoo Park$^{2}$, and Sunghoon Im$^{1,\dagger}$% <-this % stops a space
\thanks{
Manuscript received: August 18, 2025; Revised: October 17, 2025; Accepted: November 9, 2025. This paper was recommended for publication by Editor Abhinav Valada upon evaluation of the Associate Editor and Reviewers’ comments.
This research was supported by Hyundai Motor Groups, Institute of Information \& Communications Technology Planning \& Evaluation(IITP) grant funded by the Korea government(MSIT) (No. RS-2025-02219277, AI Star Fellowship Support Project(DGIST)), Basic Science Research Program through the National Research Foundation of Korea (NRF) funded by the Ministry of Education (RS-2025-25420118) and LG AI STAR Talent Development Program for Leading Large-Scale Generative AI Models in the Physical AI Domain (RS-2025-25442149) (*: Equal Contribution, $\dagger$: Corresponding Author).}% <-this % stops a space
\thanks{$^{1}$Kyumin Hwang, Wonhyeok Choi, Kiljoon Han, Wonjoon Choi, Minwoo Choi, and Sunghoon Im are with the Department of Electrical Engineering \& Computer Sciences, Daegu Gyeongbuk Institute of Science and Technology (DGIST), Daegu 42988, South Korea {\tt\small (kyumin@dgist.ac.kr; smu06117@dgist.ac.kr; kiljoon\_h@dgist.ac.kr; wjchoi@dgist.ac.kr; subminu@dgist.ac.kr; sunghoonim@dgist.ac.kr)}.
}
\thanks{$^{2}$Yongcheon Na and Minwoo Park are with the Department of Autonomous Driving Perception Technology Vanguard Team, Hyundai Motor Company, Gyeonggi 13529, South Korea {\tt\small (ycna@hyundai.com, minwoo.park@hyundai.com)}
}%
\thanks{\quad Digital Object Identifier (DOI): see top of this page.}
}

\begin{document}

\maketitle
% \thispagestyle{empty}
% \pagestyle{empty}

%%%%%%%%%%%%%%%%%%%%%%%%%%%%%%%%%%%%%%%%%%%%%%%%%%%%%%%%%%%%%%%%%%%%%%%%%%%%%%%%
\begin{abstract}
Recent foundation models demonstrate strong generalization capabilities in monocular depth estimation.
However, directly applying these models to Full Surround Monocular Depth Estimation (FSMDE) presents two major challenges: (1) high computational cost, which limits real-time performance, and (2) difficulty in estimating metric-scale depth, as these models are typically trained to predict only relative depth.
To address these limitations, we propose a novel knowledge distillation strategy that transfers robust depth knowledge from a foundation model to a lightweight FSMDE network.
Our approach leverages a hybrid regression framework combining the knowledge distillation scheme--traditionally used in classification--with a depth binning module to enhance scale consistency.
Specifically, we introduce a cross-interaction knowledge distillation scheme that distills the scale-invariant depth bin probabilities of a foundation model into the student network while guiding it to infer metric-scale depth bin centers from ground-truth depth.
Furthermore, we propose view-relational knowledge distillation, which encodes structural relationships among adjacent camera views and transfers them to enhance cross-view depth consistency.
Experiments on DDAD and nuScenes demonstrate the effectiveness of our method compared to conventional supervised methods and existing knowledge distillation approaches.
Moreover, our method achieves a favorable trade-off between performance and efficiency, meeting real-time requirements.
\end{abstract}

%%%%%%%%%%%%%%%%%%%%%%%%%%%%%%%%%%%%%%%%%%%%%%%%%%%%%%%%%%%%%%%%%%%%%%%%%%%%%%%%
\section{Introduction}
\label{sec:intro}

\begin{figure}[!t]
    \centering
    \includegraphics[width=0.9\linewidth]{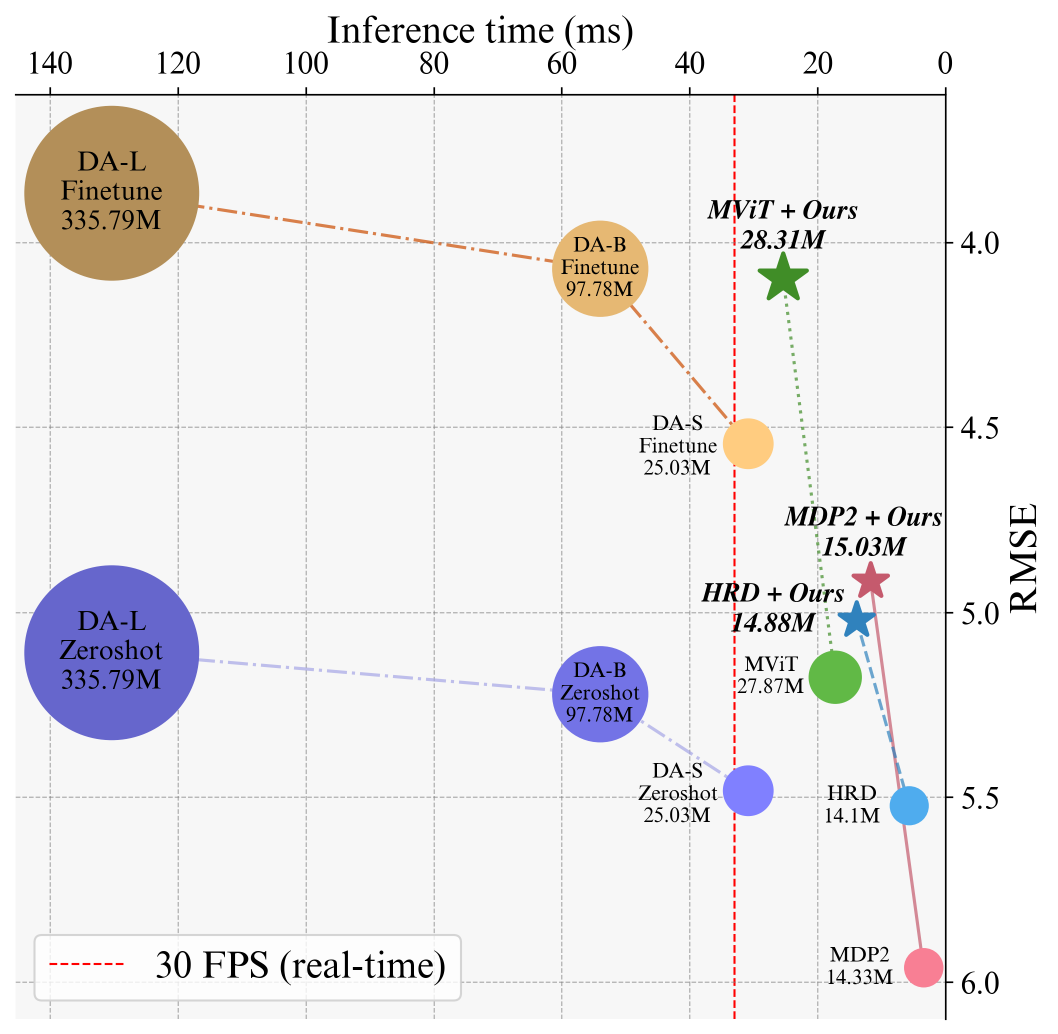}
    \caption{Inference speed vs. RMSE trade-off curve for \textbf{nuScenes} dataset (upper-right is optimal). Each circle represents a model size, and the number inside each circle indicates the number of model parameters. (DA: DepthAnything~\cite{yang2024depth}, MDP2: Monodepth2~\cite{godard2019digging}, HRD: HRDepth~\cite{lyu2021hr}, MViT: MonoViT~\cite{zhao2022monovit})}
    \label{fig:teasor}
    \vspace{-8pt}
\end{figure}

Full surround camera systems have become increasingly popular in autonomous vehicles, providing a cost-effective alternative to LiDAR-based solutions by capturing a comprehensive view of the environment.
In this context, Full Surround Monocular Depth Estimation (FSMDE)~\cite{guizilini2022full, wei2023surrounddepth, shi2023ega} has emerged as a practical and affordable solution, attracting significant research attention. 
Recent FSMDE methods have focused on developing lightweight network architectures that effectively balance computational efficiency and depth estimation accuracy, enabling reliable real-time decision-making for autonomous vehicles.

To this end, prior FSMDE approaches have employed lightweight networks trained in either self-supervised ~\cite{guizilini2022full, wei2023surrounddepth, kim2022self, schmied2023r3d3, shi2023ega} or supervised~\cite{guo2023simple} settings to efficiently learn depth by incorporating inter-view geometric consistency or spatio-temporal cues.
Meanwhile, recent advancements in foundation models~\cite{kirillov2023segment, rombach2022high, yang2024depth, yang2025depth} demonstrate remarkable generalization capabilities across various tasks, including monocular depth estimation~\cite{yang2024depth, yang2025depth}.
These large-scale models, trained on vast datasets, exhibit strong robustness in estimating relative depth across various environments.
However, applying foundation models to FSMDE for autonomous driving presents two critical challenges:
\begin{itemize}[leftmargin=17pt]
    \item[(1)] Computational cost: Foundation models are inherently large and computationally expensive, making real-time inference infeasible for autonomous driving.
    \item[(2)] Metric-scale depth estimation: Foundation models, trained on diverse datasets with varying camera intrinsics, typically produce only relative depth, making it difficult to ensure consistent metric depth across views.
\end{itemize}

\begin{figure}[t]
    \centering
    \includegraphics[width=0.9\linewidth]{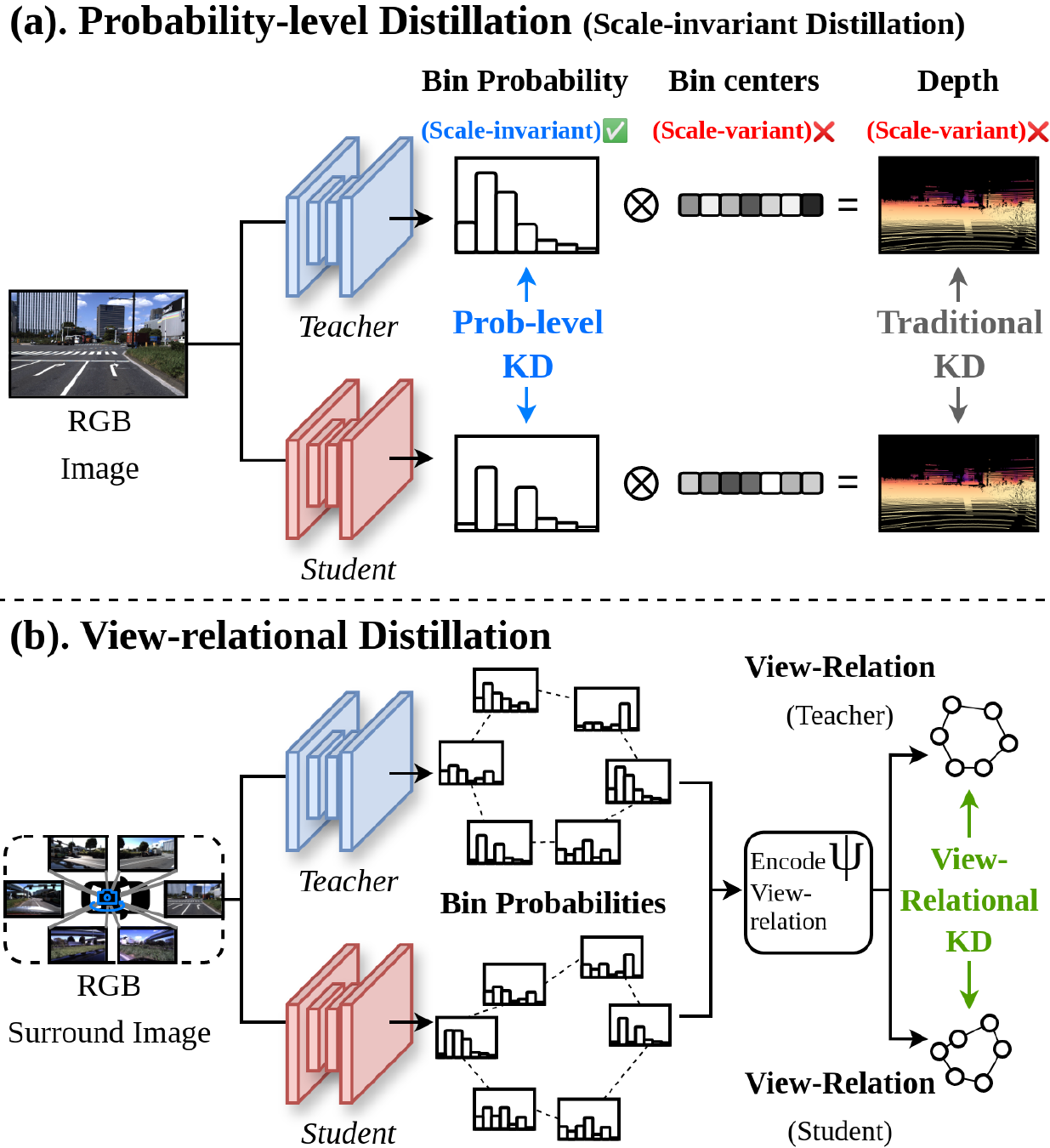}
    \caption{Conceptual illustration of our method. (a) Leveraging an effective depth binning module from supervised methods, we perform scale-invariant distillation at the probability level, avoiding the scale sensitivity of output-level distillation. (b) We use a potential function between adjacent views to distill relational information.}
    \label{fig:conceptual}
\end{figure}

To address these limitations, we propose a \textit{Cross-interaction Knowledge Distillation} (CKD) scheme that transfers robust depth information from a foundation model to a lightweight FSMDE student network.
Our method builds on the widely used hybrid regression paradigm~\cite{bhat2021adabins, bhat2022localbins, li2024binsformer}, where depth binning techniques effectively improve scale consistency in supervised Monocular Depth Estimation (MDE).
The depth binning module serves two primary functions: regressing depth bin centers to represent scale-variant depth distributions, and predicting scale-invariant depth probabilities for each pixel.
The proposed CKD distills the foundation model's bin probabilities into the student network, ensuring that it captures the foundation model's scale-invariant and generalized representation as illustrated in \figref{fig:conceptual}-(a).
% CKD distills the foundation model’s bin probabilities into the student network, ensuring it learns metric-scale depth bin centers from ground-truth depth while capturing the teacher’s generalized representation. %to capture the foundation model's generalized representation. 

Additionally, we propose a \textit{View-relational Knowledge Distillation} (VRKD), which transfers the inter-camera relationships learned by the teacher model to the student model.
It allows the student to leverage spatial information from adjacent camera views in FSMDE.
Conventional FSMDE methods typically refine metric depth estimation by enforcing geometric constraints or incorporating spatio-temporal cues across the full surround camera system.
In contrast, our approach distills the structural relationships between cameras into the student network at the probability level as shown in \figref{fig:conceptual}-(b).
Inspired by relational knowledge distillation~\cite{park2019relational}, our framework encodes depth distribution relationships between adjacent views using a potential function that measures the relational energy across the $N$-camera-view system.
This enables our framework to transfer the teacher’s view-relational structure to the student network, enhancing multi-view depth consistency.

Through extensive experiments, we demonstrate that our method achieves an average improvement of 5.88\% on DDAD and 11.87\% on nuScenes compared to conventional supervised MDE.
Additionally, our approach offers $5.13$ -- $11.14\times$ faster inference speed compared to the teacher foundation model, enabling real-time inference on practical autonomous driving applications as shown in \figref{fig:teasor}.
We also compare our method with existing knowledge distillation methods~\cite{hinton2015distilling, romero2014fitnets, zagoruyko2016paying, tung2019similarity}, showing that our method significantly outperforms these techniques in the FSMDE training scenario.

\noindent Our main contributions are summarized as follows:
\begin{itemize}[leftmargin=17pt]
    \item We propose a \textit{Cross-interaction Knowledge Distillation} (CKD) scheme that transfers a scale-invariant depth knowledge from a foundation model to a lightweight student network.
    \item We present a \textit{View-relational Knowledge Distillation} (VRKD) that effectively distills inter-camera relationships from the teacher to the student network.
    \item We validate the effectiveness of our method by surpassing both depth-supervised approaches and existing distillation methods across two FSMDE datasets.
\end{itemize}

\section{Related Work}
\label{sec:related_work}

\subsection{Monocular Depth Estimation}
% (MDE 초기) Eigen14, DORN18, BTS19, DPT21, Adabins21, Localbins22, GLPDepth22, BinFormer23, PixelFormer23
Since~\cite{eigen2014depth} introduced deep learning-based monocular depth estimation, the field has seen active progress~\cite{xu2018structured, lee2019big, choi2023depth}.
Subsequent advancements include ViT-based architectures for effective global context embedding~\cite{ranftl2021vision, kim2022global} and methods leveraging geometric priors, such as ground planes or normal maps~\cite{shao2023nddepth, zhao2021camera}. 
Pioneering supervised methods like DORN~\cite{fu2018deep} discretized depth estimation into a classification task with fixed intervals. 
Subsequent works advanced this by introducing adaptive binning, where methods like AdaBins~\cite{bhat2021adabins} and LocalBins~\cite{bhat2022localbins} dynamically predict depth distributions at the global and per-pixel levels, respectively, to better suit scene-specific content.
More recently, ZoeDepth~\cite{bhat2023zoedepth} introduced the MetricBins module to enable the joint estimation of relative and metric depth by iteratively refining bin centers.
Moreover, depth foundation models such as Depth Anything~\cite{yang2024depth} have emerged, achieving robust generalization across diverse environments.
% More recently, ZoeDepth~\cite{bhat2023zoedepth} introduces the MetricBins module, which iteratively refines bin centers to simultaneously estimate relative and metric depth, highlighting a shift toward metric monocular depth estimation.

\subsection{Full Surround Monocular Depth Estimation}

Numerous self-supervised monocular depth estimation methods~\cite{godard2019digging, lyu2021hr, zhao2022monovit, choi2025self, bae2023study} have achieved impressive performance without relying on explicit ground-truth supervision.
Building upon these advancements, FSM~\cite{guizilini2022full} pioneers Full Surround Monocular Depth Estimation (FSMDE), offering a cost-efficient approach to multi-camera depth estimation for autonomous driving by incorporating multi-camera spatio-temporal context to reconstruct scale-aware depth.
Following FSM, several methods~\cite{kim2022self, wei2023surrounddepth} leverage cross-view and temporal information to enhance depth estimation.
For instance, these works employ diverse strategies such as cross-view self-attention~\cite{wei2023surrounddepth} and volumetric feature fusion~\cite{kim2022self}.

\subsection{Knowledge Distillation}
Knowledge Distillation (KD)~\cite{hinton2015distilling} has emerged as an effective technique for model compression and knowledge transfer.
The core idea is to guide the student to mimic the teacher’s knowledge through various mechanisms, such as aligning predictions~\cite{chen2017learning}, transferring intermediate features~\cite{romero2014fitnets, zagoruyko2016paying}, or capturing structural relationships between features~\cite{park2019relational, tung2019similarity}.
In computer vision, KD has been widely applied to tasks such as image classification~\cite{li2017learning, peng2019few}, object detection~\cite{chen2017learning}.
For depth estimation, \cite{pilzer2019refine} enhanced unsupervised monocular methods using KD with an error correction mechanism, while \cite{wang2021knowledge} introduced a feature-based KD framework for mobile devices.
% For depth estimation, \cite{pilzer2019refine} improves unsupervised monocular depth by combining KD with an error correction mechanism.
% \cite{wang2021knowledge} proposes a feature-based KD framework for efficient depth estimation on mobile devices.

% \subsection{Depth Foundation Model}

% Recent research has focused on depth foundation models, pre-trained across diverse domains and adaptable to various downstream tasks, to improve zero-shot transfer performance.
% Early work, MiDaS~\cite{ranftl2020towards}, introduces a scale- and shift-invariant loss along with a data mixing strategy to train on heterogeneous datasets. DPT~\cite{ranftl2021vision} further improves transferability by replacing the CNN backbone with ViTs~\cite{dosovitskiy2020image}.
% More recently, DepthAnything~\cite{yang2024depth, yang2025depth} and Marigold~\cite{ke2024repurposing} leverage pre-trained models for zero-shot knowledge transfer, DINOv2~\cite{oquab2023dinov2} and Stable Diffusion~\cite{rombach2022high}, respectively.
% To enhance metric depth estimation, UniDepth~\cite{piccinelli2024unidepth} incorporates dense camera representations, while DepthPro~\cite{bochkovskii2024depth} introduces a multi-scale ViT-based framework requiring no additional metadata.
% Some approaches address geometric domain gaps by integrating camera information, such as applying canonical camera space transformations to resolve metric ambiguity~\cite{hu2024metric3d} and jointly encoding camera parameters and image features for scale-aware depth estimation~\cite{guizilini2023towards}.
\section{Method}
\label{sec:method}

\begin{figure*}[!t]
    \centering
    \includegraphics[width=0.8\linewidth]{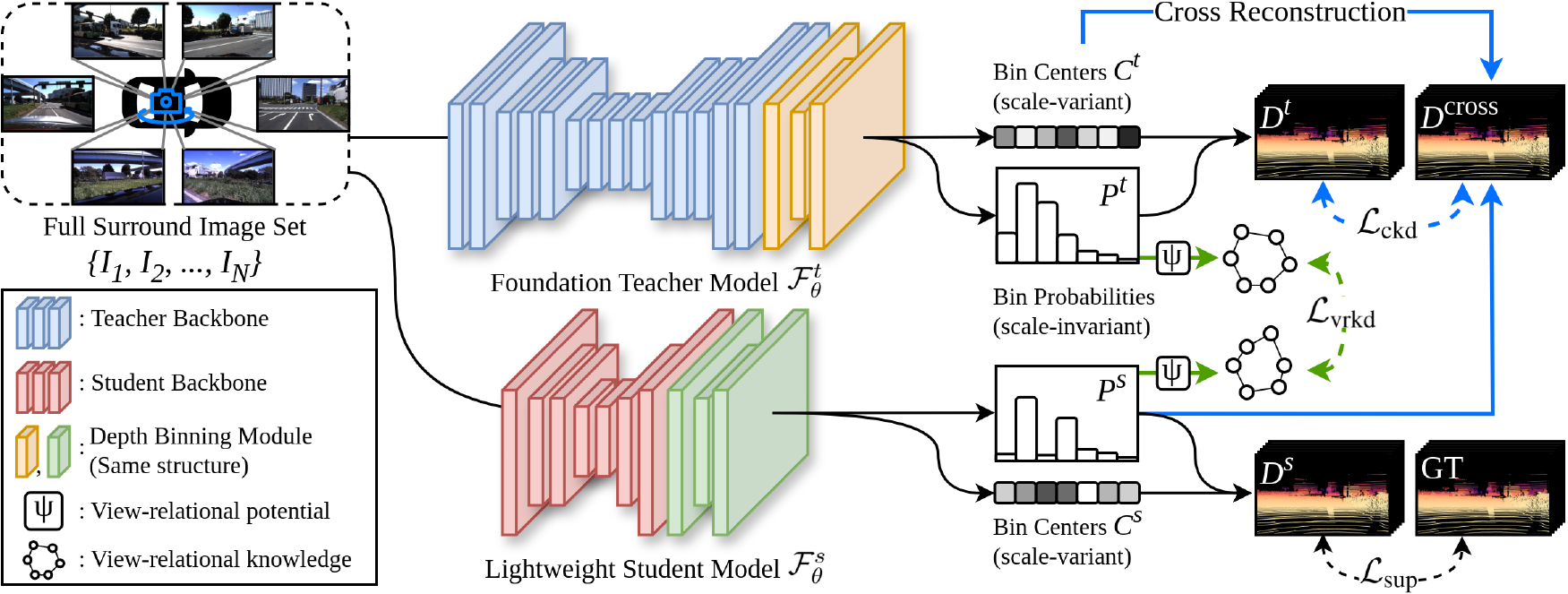}
    \caption{Illustration of the proposed knowledge distillation schemes. Our method leverages a depth binning module with the same architecture as the teacher model, enabling effective knowledge distillation at the scale-invariant depth bin probability level.}
    \label{fig:cross}
    % \vspace{-10pt}
\end{figure*}

\subsection{Problem Definition}
\label{subsec:problem_definition}

% \begin{comment}
The task of full surround monocular depth estimation aims to predict the metric depth maps from given a set of input surrounding samples $\mathbf{I} = \{I_i\}_{i=1}^{N}$ and the corresponding ground truth depth maps $\mathbf{D} = \{D_i\}_{i=1}^{N}$, where the $N$ is the number of surround cameras.
The objective of our method is to improve the depth estimation performance across the surround camera views by distilling the knowledge of a teacher foundation model to a student model.

% % bins module로 한정지을 때
% We assume that the teacher model $\mathcal{F}_{\theta}^{t}$ and student model $\mathcal{F}_{\theta}^{s}$ adopt the same structure of depth binning module for hybrid regression~\cite{bhat2021adabins, bhat2022localbins}.
% Depth Binning method는 픽셀별 depth 값을 직접 예측하는 대신, depth 값을 $B$개의 bin 으로 나눈다.
% 동시에, 모델 $\mathcal{F}$는 each bin에 대한 depth-bin-center $C = \{c_1, c_2, \cdots, c_B\}, c_i \in \mathbb{R}^{C \times H \times W}$와 corresponding probablities $P = \{p_1, p_2, \cdots, p_B\}$를 pixel-wise하게 예측한뒤 이를 이용해 각 뷰에 대해서 최종 depth $\hat{D}$를 계산한다 as follows:
% The model $\mathcal{F}_{\theta}$ jointly predicts depth bin centers $C$ and probabilities $P$ from the input image $I$:
% \begin{align}
% C, P = \mathcal{F}_{\theta}(I), ~~\text{where}~C = \{c_k\}_{k=1}^B,~P = \{p_k\}_{k=1}^B, \quad c_k, p_k \in \mathbb{R}^{H \times W},
% \end{align}
% Then, the final per-pixel depth estimation $\hat{D}$ is reconstructed using a weighted sum:
% \begin{align}
% \label{eq:depth_recon}
% \hat{D}[u,v] = \sum_{k=1}^{B} c_{k}[u, v] \cdot p_k[u,v], ~\forall u \in \{1,\cdots,H\},~ v \in \{1,\cdots,W\}.
% \end{align}

As aforementioned, we assume that the teacher model $\mathcal{F}_{\theta}^{t}$ and the student model $\mathcal{F}_{\theta}^{s}$ adopt the same depth binning module structure for hybrid regression~\cite{bhat2021adabins, bhat2022localbins}.  
Rather than directly predicting per-pixel depth values, the depth binning method discretizes depth into $B$ bins.  
Simultaneously, the model $\mathcal{F}$ jointly predicts the pixel-wise depth bin centers $C$, and the corresponding probabilities $P$ as follows:
% \begin{align}
% % P, C = \mathcal{F}_{\theta}(I),
% C, P = \mathcal{F}_{\theta}(I), ~~\text{where}~C = \{c_k\}_{k=1}^B,~P = \{p_k\}_{k=1}^B, \quad c_k, p_k \in \mathbb{R}^{H \times W},
% \end{align}
\begin{equation}
\begin{gathered}
% P, C = \mathcal{F}_{\theta}(I),
C, P = \mathcal{F}_{\theta}(I), \\
~~\text{where}~C = \{c_k\}_{k=1}^B,~P = \{p_k\}_{k=1}^B, 
\end{gathered}
\end{equation}

where $H$ and $W$ denote height and width of an image $I$, respectively, and $c_k, p_k \in \mathbb{R}^{H \times W}$.
By combining these depth bin centers and probabilities, the final depth estimation $\hat{D}$ for each view is then computed as follows:
% \begin{align}
% \begin{split}
% \label{eq:depth_recon}
% \hat{D}[u,v] = \sum_{k=1}^{B} c_{k}[u, v] \cdot p_k[u,v], ~\forall u \in \{1,\cdots,H\},~ v \in \{1,\cdots,W\},
% \end{split}
% \end{align}
\begin{equation}
\label{eq:depth_recon}
\begin{gathered}
    \hat{D}[u,v] = \sum_{k=1}^{B} c_{k}[u, v] \cdot p_k[u,v], \\
    ~\forall u \in \{1,\cdots,H\},~ v \in \{1,\cdots,W\},
\end{gathered}
\end{equation}

where $u$ and $v$ denote the indices corresponding to the width $W$ and height $H$ of the image, respectively.
Note that each depth bin center $c_k \in C$ is adaptively determined for each sample through the network's prediction.
In conventional supervised methods, the student model predicts the depth bin centers $C^s$ and the corresponding probabilities $P^s$, which are then used to compute the student's depth output $\hat{D}^s$ via the above formulation.
The error between $\hat{D}^s$ and the ground-truth depth $D$ is minimized using a specific loss term $L_{\text{depth}}$ (\textit{e.g.,} L1, SiLog~\cite{eigen2014depth}) as follows:
\begin{align}
\begin{split}
\label{eq:sup}
\mathcal{L}_{\text{sup}} = L_{\text{depth}}(\hat{D}^{s}, D).
\end{split}
\end{align}

% Note that $C$ is the 네트워크의 예측을 통해 샘플마다 적합한 스케일로 구해진 depth-bin-center이다.
% 기존의 supervised method에서는 주로 student 예측한 depth-bin-centers $C^s$와 corresponding probability $P^s$를 위의 수식을 통해 student's depth output $\hat{D}^s$을 구하고, gt depth $D$와의 특정 loss term $L_{\text{depth}}$ (e.g., L1, SiLog~\cite{})으로 error를 minimize한다 as follows:
% teacher와 student가 각각 예측한 depth-bin-centers $C^t, C^s$ 와 corresponding probability $P^t, P^s$ 를 위의 수식을 통해 $D^t, D^s$ 로 계산한 후 기존의 방법
% depth-bin-center set $C$는 strictly increasing sequence이고 sample의 depth distribution에 맞게 adaptively bin-center를 선택한다.

\subsection{Cross-interaction Knowledge Distillation}
\label{subsec:ckd}

The network should infer the absolute scale (\textit{i.e.,} metric scale) depth bin centers $C$ based on the depth distribution of each sample.  
% On the other hand, depth bin centers와의 weighted sum으로 depth를 추정하는데 이용되는 the probability $P$ 는 실제 스케일이 고려되지 않은 상대적 깊이에 대한 확률 정보를 담고 있다.
On the other hand, the probability $P$, which is used to estimate depth through the weighted sum of depth bin centers, represents the probability distribution of relative depth without considering the actual scale.  
% 반면 probability $P$는 실제 스케일이 고려되지 않은 상대적인 깊이에 대한 direct 정보를 담고 있다.
The main idea of our method is to improve overall depth accuracy by distilling the probability $P^t$ of the foundation model--which encapsulates robust and scale-invariant information--into the student’s probability $P^s$, while simultaneously training the student's depth bin centers $C^s$ to align with the actual metric scale of ground truth depth.  
% 따라서 우리의 방법의 main idea는 강건한 상대적 깊이를 추정할 수 있는 foundation model의 probability $P^t$를 학생 네트워크의 probability $P^s$로 distillation함과 동시에, 학생의 depth-bin-centers $C^{s}$를 실제 metric depth에 맞도록 학습함으로써 성능증가를 도모하는 것이다.
% 이를 위해, 우리는 teacher와 student의 depth-bin-centers와 probabilities를 활용하여 systematically 위의 목적을 달성하는 cross-interaction knowledge distillation을 제안한다.

To achieve this, we propose a \textbf{\textit{cross-interaction knowledge distillation}} scheme that systematically transfers the robust representation of the teacher model’s relative depth to the student network by leveraging bin probabilities and bin centers from the teacher and student models, respectively.  
The overall training mechanism is depicted in \figref{fig:cross}.
We first obtain the probabilities and depth bin centers of both teacher network  $\mathcal{F}_{\theta}^t$ and student network $\mathcal{F}_{\theta}^s$ as follows:
\begin{align}
\begin{split}
\label{eq:inference}
C^t, P^t = \mathcal{F}_{\theta}^{t}(I),\quad C^s, P^s = \mathcal{F}_{\theta}^{s}(I).
\end{split}
\end{align}

To transfer the probability distribution from the teacher network to the student network, we employ a loss function $\mathcal{L}_{\text{ckd}}$ that reconstructs the teacher’s depth prediction using the teacher’s bin centers and the student’s probability distribution, formulated as follows:
% \begin{align}
% \begin{split}
% \label{equ:cross1}
% \mathcal{L}_{\text{ckd}} = L_{\text{depth}}(\hat{D}^{\text{cross}}, \hat{D}^{t}), ~~\text{where}~\hat{D}^{\text{cross}}[u,v] = \sum_{k=1}^{B} c_{k}^t[u, v] \cdot p_k^s[u,v],\\
% \end{split}
% \end{align}
\begin{equation}
\label{equ:cross1}
\begin{gathered}
\mathcal{L}_{\text{ckd}} = L_{\text{depth}}(\hat{D}^{\text{cross}}, \hat{D}^{t}), \\
\text{where} \quad \hat{D}^{\text{cross}}[u,v] = \sum_{k=1}^{B} c_{k}^t[u, v] \cdot p_k^s[u,v]
\end{gathered}
\end{equation}
where $c_k^t \in C^t,~p_k^s\in P^s$, and $\hat{D}^{t}$ indicate the teacher's depth prediction, respectively.
The objective of this loss is to encourage the student network to learn the teacher's probability, while student depth bin centers are guided by \equref{eq:sup}.

\subsection{View-relational Knowledge Distillation}
\label{subsec:viewkd}
Existing full surround self-supervised monocular depth estimation approaches~\cite{guizilini2022full, wei2023surrounddepth,kim2022self,shi2023ega} commonly leverage the geometric constraint from spatio-temporal information from each camera coordinate.
On the other hand, in a supervised scenario where the absolute scale depth (\textit{i.e.,} metric depth for each viewpoint) is given as ground truth, we propose a learning-based method, \textit{\textbf{view-relational knowledge distillation}}, which allows the structural knowledge between camera views to be distilled from the teacher model to the student model.
This method aims to transfer the teacher's structural knowledge using mutual relationships between depth distributions from each adjacent camera view.
Similar to \cite{park2019relational}, given teacher's depth bin probabilities across all cameras $\{\hat{P}_i^t\}_{i=1}^{N}$ and student depth bin probabilities $\{\hat{P}_i^s\}_{i=1}^{N}$ through \equref{eq:inference}, we first encode the internal pair-wise relations of the teacher's and student's depth bin probabilities using the relational potential function $\psi$ as follows:
% we first encode the pair-wise relation of the teacher and student probabilities by using the relational potential function $\psi$ as follows:
% \begin{align}
% \begin{split}
% \label{eq:encode_rel}
% \Gamma_{(i,j)}^t = \psi(\hat{P}_i^t, \hat{P}_j^t),~\Gamma_{(i,j)}^s = \psi(\hat{P}_i^s, \hat{P}_j^s),~~\forall
% \psi(P_i, P_j) = \frac{1}{\mu} \|P_i - P_j\|_2,
% \end{split}
% \end{align}
\begin{equation}
\label{eq:encode_rel}    
\begin{gathered}
\Gamma_{(i,j)}^t = \psi(\hat{P}_i^t, \hat{P}_j^t),~\Gamma_{(i,j)}^s = \psi(\hat{P}_i^s, \hat{P}_j^s), \\
~~\forall \psi(P_i, P_j) = \frac{1}{\mu} \|P_i - P_j\|_2,    
\end{gathered}
\end{equation}

where $i$ and $j$ are the camera indices, and $\mu$ is the normalization factor for distance, respectively.

To pair each camera $i$ with its adjacent camera~$j$, which shares overlapping fields of view, we define the adjacent camera pair set $(i,j)$ within the set $\mathcal{A}= \{(1,2), (2,3), \cdots, (N,1)\}$, defined cyclically.
The student network is trained using the view-relational knowledge distillation loss $\mathcal{L}_{\text{vrkd}}$, which encourages the student to mimic the teacher's camera-wise relations.
Specifically, the relation between adjacent camera views $\Gamma_{i,j}^s$ in the student network is encouraged to approximate $\Gamma_{i,j}^t$ from the teacher using the Huber loss~\cite{huber1992robust} $L_{\text{huber}}$, as follows:
\begin{align}
\begin{split}
\mathcal{L}_{\text{vrkd}} = \sum_{(i,j) \in \mathcal{A}} L_{\text{huber}}(\Gamma_{i,j}^t, \Gamma_{i,j}^s).
\end{split}
\end{align}

This loss distills the relational information of the adjacent cameras' probabilities--including the scale-invariant depth knowledge--to the student, thereby enabling the student to learn the relationships between adjacent camera views.
In the end, the final loss term of our method consists of a weighted sum of the supervised loss term $\mathcal{L}_{\text{sup}}$, the cross-interaction knowledge distillation loss $\mathcal{L}_{\text{ckd}}$ introduced in \secref{subsec:ckd}, and $\mathcal{L}_{\text{vrkd}}$ as follows:
\begin{align}
\begin{split}
\mathcal{L}_{\text{total}} = \mathcal{L}_{\text{sup}} + \lambda_{\text{ckd}} \cdot \mathcal{L}_{\text{ckd}} + \lambda_{\text{vrkd}} \cdot \mathcal{L}_{\text{vrkd}},
\end{split}
\end{align}
where $\lambda_{\text{ckd}}$ and $\lambda_{\text{vrkd}}$ are the weight balancing parameters for each loss terms.
\begin{figure*}[t]
    \centering
    \includegraphics[width=0.9\linewidth]{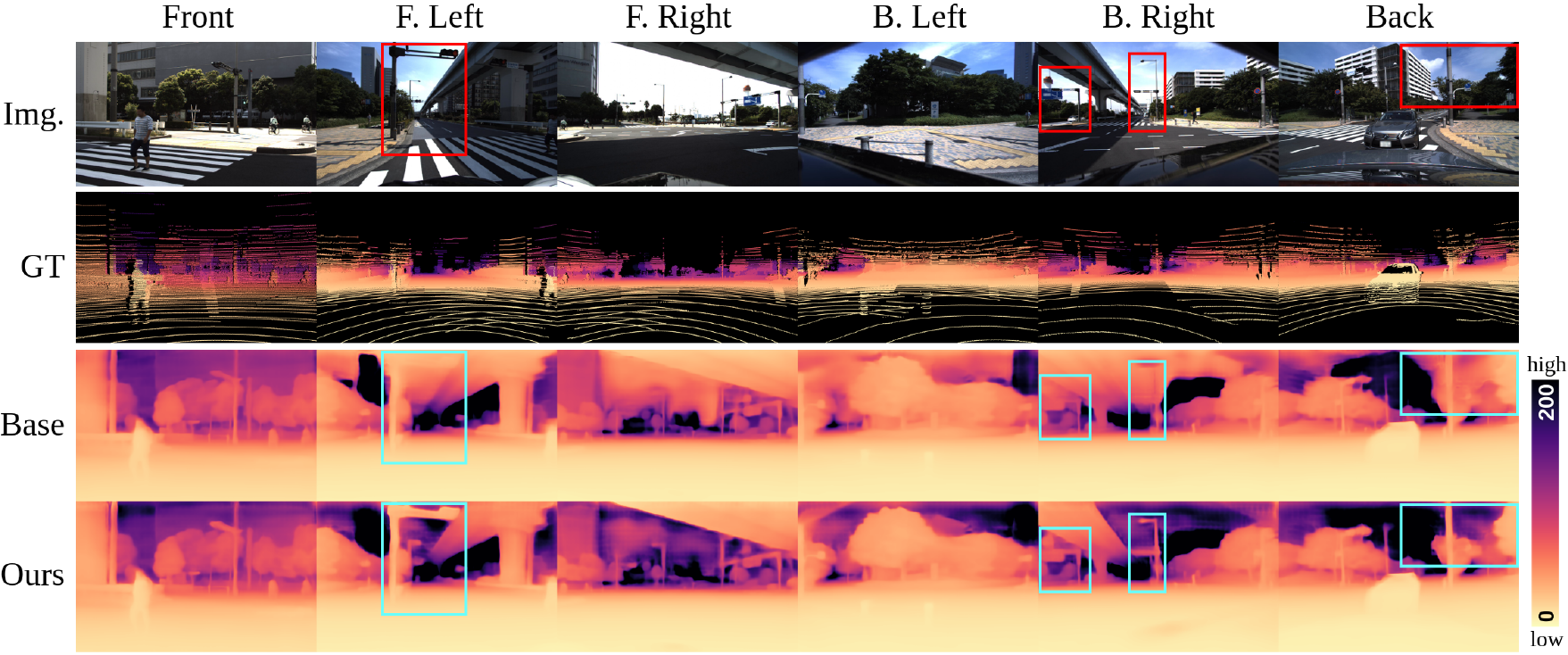}
    % \vspace{-10pt}
    \caption{Qualitative results of fine-tuned Monodepth2 (denoted as Base) and Monodepth2 + Ours (denoted as Ours) on \textbf{DDAD} dataset.}
    % \vspace{-10pt}
    
    %The second and third rows of each sample show the depth prediction, and the last two rows present the error map. We report the dense depth visualization in the appendix.}
    \label{fig:qualitative_ddad}
\end{figure*}
\begin{table*}[t]
    \caption{Evaluation results on \textbf{DDAD} dataset. For the binning module for students, we adopt the MetricBins (MB), which is the same architecture as the teacher's (\textit{i.e., DepthAnything}) binning module. $\Delta_\mathcal{T}$ denotes the relative performance increment from each baseline. Unlike supervised FSMDE, we apply median scaling for self-supervised methods.}
    \centering
    \resizebox{0.99\linewidth}{!}{
    \label{tab:ddad}
    \begin{tabular}{clcccccccr|rr}
        \toprule
        % Methods & Abs Rel $\downarrow$ & Sq Rel $\downarrow$ & RMSE $\downarrow$ & RMSE log $\downarrow$ & $\delta < 1.25 \uparrow$ & $\delta < 1.25^2 \uparrow$ & $\delta < 1.25^3 \uparrow$ & $\Delta_\mathcal{T} (\%) \uparrow$ & Params. $\downarrow$ & Latency $\downarrow$ \\
        % & Methods & Abs Rel \textcolor{red}{$\downarrow$} & Sq Rel \textcolor{red}{$\downarrow$} & RMSE \textcolor{red}{$\downarrow$} & RMSE log \textcolor{red}{$\downarrow$} & $\delta < 1.25$ \textcolor{red}{$\uparrow$} & $\delta < 1.25^2$  \textcolor{red}{$\uparrow$} & $\delta < 1.25^3$  \textcolor{red}{$\uparrow$} & $\Delta_\mathcal{T} (\%)$ \textcolor{red}{$\uparrow$} & Params. $\downarrow$ & Latency $\downarrow$ \\
        & Methods & Abs Rel $\downarrow$ & Sq Rel $\downarrow$ & RMSE $\downarrow$ & RMSE log $\downarrow$ & $\delta < 1.25$ $\uparrow$ & $\delta < 1.25^2$ $\uparrow$ & $\delta < 1.25^3$ $\uparrow$ & $\Delta_\mathcal{T} (\%)$ $\uparrow$ & Params. $\downarrow$ & Latency $\downarrow$ \\
        \midrule
        % DepthAnything-S (zero-shot) & 0.287 & 3.571 & 12.050 & 0.634 & 0.577 & 0.810 & 0.898 & & 25.03 & \\
        % DepthAnything-B (zero-shot) & 0.281 & 3.412 & 12.016 & 0.652 & 0.580 & 0.816 & 0.898 & & 97.78 & \\
        & \textit{DepthAnything (zero-shot)} & \textit{0.270} & \textit{3.291} & \textit{11.866} & \textit{0.621} & \textit{0.601} & \textit{0.826} & \textit{0.904} & - & \textit{335.79} M & \textit{142.26} ms \\
        % DepthAnything-S (finetuned) & 0.160 & 2.178 & 10.284 & 0.248 & 0.792 & 0.923 & 0.963 & & 25.03 & \\
        % DepthAnything-B (finetuned) & 0.148 & 1.995 & 9.860 & 0.237 & 0.815 & 0.930 & 0.967 & & 97.78 & \\
        & \textit{DepthAnything (finetuned)} & \textit{0.140} & \textit{1.866} & \textit{~9.475} & \textit{0.228} & \textit{0.831} & \textit{0.935} & \textit{0.969} & - & \textit{335.79} M & \textit{142.26} ms \\
        \midrule
        \multirow{6}{*}{\rotatebox[origin=c]{90}{Self-supervised}} & Monodepth2~\cite{godard2019digging} & 0.217 & 3.641 & 12.962 & 0.323 & 0.699 & 0.877 & 0.939 & - & 14.33 M & 3.75 ms \\
        & PackNet-SfM~\cite{guizilini20203d} & 0.234 & 3.802 & 13.253 & 0.331 & 0.672 & 0.860 & 0.931 & - & 128.29 M & 63.08 ms \\
        & FSM~\cite{guizilini2022full} & 0.202 & - & - & - & - & - & - & - & 14.33 M & 3.75 ms \\
        & SurroundDepth~\cite{wei2023surrounddepth} & 0.200 & 3.392 & 12.270 & 0.301 & 0.740 & 0.894 & 0.947 & - & 59.56 M & 11.79 ms \\
        & EGA-Depth-LR~\cite{shi2023ega} & 0.195 & 3.211 & 12.117 & 0.297 & 0.743 & 0.896 & 0.947 & - & not public & not public \\
        & EGA-Depth-MR~\cite{shi2023ega} & \textbf{0.191} & \textbf{3.126} & \textbf{11.922} & \textbf{0.290} & \textbf{0.747} & \textbf{0.901} & \textbf{0.950} & - & not public & not public \\
        \midrule
        \multirow{9}{*}{\rotatebox[origin=c]{90}{Supervised}} & Monodepth2 & 0.200 & 3.087 & 12.849 & 0.323 & 0.679 & 0.861 & 0.932 & 0.00 & 14.33 M & 3.75 ms \\
        & Monodepth2 + MB & 0.195 & 3.016 & 12.727 & 0.322 & 0.686 & 0.865 & 0.932 & $+$1.08 & 15.03 M & 12.77 ms \\
        & Monodepth2 + MB + Ours & \textbf{0.191} & \textbf{2.865} & \textbf{12.134} & \textbf{0.300} & \textbf{0.710} & \textbf{0.881} & \textbf{0.943} & \textbf{$+$4.64} & 15.03 M & 12.77 ms \\
        \cmidrule{2-12}
        & HRDepth & 0.196 & 2.955 & 12.428 & 0.304 & 0.699 & 0.875 & 0.940 & 0.00 & 14.10 M & 6.19 ms \\
        & HRDepth + MB & 0.194 & 2.961 & 12.399 & 0.303 & 0.700 & 0.877 & 0.942 & $+$0.28 & 14.80 M & 15.18 ms\\
        & HRDepth + MB + Ours & \textbf{0.183} & \textbf{2.684} & \textbf{11.578} & \textbf{0.283} & \textbf{0.736} & \textbf{0.896} & \textbf{0.950} & \textbf{$+$5.47} & 14.80 M & 15.18 ms \\
        % \midrule
        \cmidrule{2-12}
        & MonoViT & 0.179 & 2.602 & 11.777 & 0.287 & 0.725 & 0.892 & 0.949 & 0.00 & 27.87 M & 18.83 ms \\
        & MonoViT + MB & 0.178 & 2.585 & 11.757 & 0.286 & 0.728 & 0.893 & 0.950 & $+$0.34 & 28.31 M & 27.71 ms \\
        & MonoViT + MB + Ours & \textbf{0.166} & \textbf{2.260} & \textbf{10.483} & \textbf{0.256} & \textbf{0.773} & \textbf{0.916} & \textbf{0.961} & \textbf{$+$7.54} & 28.31 M & 27.71 ms \\
        \bottomrule
    \end{tabular}}
\end{table*}

\section{Experiments}
\label{sec:experiments}

\paragraph{Datasets} We evaluate our method on the \textit{DDAD}~\cite{guizilini20203d} and \textit{nuScenes}~\cite{caesar2020nuscenes} datasets.
DDAD captures diverse urban environments in the US and Japan using six cameras and a Luminar-H2 sensor, covering a full 360-degree field of view. 
It contains 73,914 training images (12,319 per camera) and 23,700 validation images (3,950 per camera).
With a maximum range of 250m and only 20\% overlap between adjacent cameras, DDAD simulates realistic full surround autonomous driving conditions.
The nuScenes dataset, widely used in autonomous driving research, is collected from urban scenes across the US and Singapore, utilizing six surrounding cameras and a LiDAR sensor. It includes 120,576 training images (20,096 per camera) and 36,114 validation images (6,019 per camera).
The camera setup has at most 10\% overlap between adjacent views, posing challenges for cross-view consistency and multi-view depth estimation.
% Its diverse environments and weather conditions make it well-suited for evaluating generalization performance~\cite{saunders2023self, gasperini2023robust}.

\paragraph{Baselines}
We adopt three representative Monocular Depth Estimation (MDE) frameworks as student networks, including Monodepth2~\cite{godard2019digging}, HRDepth~\cite{lyu2021hr}, and MonoViT~\cite{zhao2022monovit}, which are widely used in MDE research. 
HRDepth and MonoViT, with their enhanced encoder-decoder architectures built on Monodepth2, are particularly suitable for evaluating the effectiveness of our proposed knowledge distillation as the student network size increases.
Notably, FSM~\cite{guizilini2022full}, SurroundDepth~\cite{wei2023surrounddepth}, and CVCDepth~\cite{ding2024towards} share the same Monodepth2 backbone and are trained in a self-supervised manner. Due to their architectural relevance and prevalence in recent literature, we select Monodepth2, HRDepth, and MonoViT as our backbone networks.
For the teacher network, we use DepthAnything~\cite{yang2024depth}, a highly generalizable model capable of inferring depth maps across diverse environments.
As described in \secref{sec:method}, we incorporate a MetricBins (MB) module---identical to the teacher's binning module---into each student network to leverage the bin probability-level representation from the teacher.
% The MB module receives fused features from the student's depth decoder, combining relative depth and multi-scale decoder features. 
All [baseline + MB] configurations use 64 bins, with a bin embedding dimension of 128.

\paragraph{Implementation details} We follow the training and evaluation protocols of state-of-the-art FSMDE methods~\cite{guizilini2022full,wei2023surrounddepth,kim2022self} to ensure fair comparisons between our approach and conventional supervised methods.
For the DDAD dataset, we set the maximum depth to 200m to reflect the long-range capability and apply the self-occlusion masks provided by SurroundDepth~\cite{wei2023surrounddepth} to exclude regions occluded by the vehicle. 
% Additionally, we use the self-occlusion masks provided by SurroundDepth~\cite{wei2023surrounddepth} to filter occluded regions, which are occluded by the vehicle itself due to the cameras being mounted inside the vehicle.
% To ensure a fair comparison, we used the self-occlusion masks provided by SurroundDepth~\cite{wei2023surrounddepth} to filter these occlusion regions.
For nuScenes, no self-occlusion mask is needed, and the maximum depth is set to 80m.
Training image resolutions are set to $640 \times 384$ for DDAD and $640 \times 352$ for nuScenes, and median scaling is not applied during evaluation.
% Additionally, since our method does not use the temporal context, we use only the current time-stamp input images from whole camera views.
Each model is trained for 5 epochs with a batch size of 12.
We follow the hyperparameter setup of supervised training and the MetricBins module of DepthAnything~\cite{yang2024depth}.
Weight balancing parameters for our method are set to $\lambda_{\text{ckd}}=0.1$ and $\lambda_{\text{vrkd}}=1.0$.
%for our method as $0.1$ and $1.0$, respectively.
The teacher model (DepthAnything) uses pre-trained weights from the outdoor datasets, while all student models are initialized with pre-trained weights from KITTI~\cite{geiger2013vision} to accelerate training and improve convergence.
We present our results following the standard evaluation metrics introduced by \cite{eigen2014depth}.
For evaluating inference time, we conduct measurements using a single NVIDIA RTX A6000.

\begin{table*}[t]
    \caption{Evaluation results on \textbf{nuScenes} dataset. $\Delta_\mathcal{T}$ denotes the relative performance increment from each baseline.} 
    %For the binning module for students, we adopt the MetricBins (MB), which is the same architecture as the teacher's (\textit{i.e., DepthAnything}) binning module. $\Delta_\mathcal{T}$ denotes the relative performance increment from each baseline.}
    \label{tab:nuscenes}
    \centering
    \resizebox{0.99\linewidth}{!}{
    \begin{tabular}{clcccccccr|rr}
        \toprule
        % & Methods & Abs Rel \textcolor{red}{$\downarrow$} & Sq Rel \textcolor{red}{$\downarrow$} & RMSE \textcolor{red}{$\downarrow$} & RMSE log \textcolor{red}{$\downarrow$} & $\delta < 1.25$ \textcolor{red}{$\uparrow$} & $\delta < 1.25^2$ \textcolor{red}{$\uparrow$} & $\delta < 1.25^3$ \textcolor{red}{$\uparrow$} & $\Delta_\mathcal{T} (\%)$ \textcolor{red}{$\uparrow$} & Params. $\downarrow$ & Latency $\downarrow$ \\
        & Methods & Abs Rel $\downarrow$ & Sq Rel $\downarrow$ & RMSE $\downarrow$ & RMSE log $\downarrow$ & $\delta < 1.25$ $\uparrow$ & $\delta < 1.25^2$ $\uparrow$ & $\delta < 1.25^3$ $\uparrow$ & $\Delta_\mathcal{T} (\%)$ $\uparrow$ & Params. $\downarrow$ & Latency $\downarrow$ \\
        % Methods & Abs Rel $\downarrow$ & Sq Rel $\downarrow$ & RMSE $\downarrow$ & RMSE log $\downarrow$ & $\delta < 1.25 \uparrow$ & $\delta < 1.25^2 \uparrow$ & $\delta < 1.25^3 \uparrow$ & $\Delta_\mathcal{T} (\%) \uparrow$ & Params. $\downarrow$ & Latency $\downarrow$ \\
        \midrule
        & \textit{DepthAnything (zero-shot)} & \textit{0.208} & \textit{1.324} & \textit{5.108} & \textit{0.280} & \textit{0.735} & \textit{0.901} & \textit{0.951} & - & \textit{335.79} M & \textit{130.41} ms \\
        & \textit{DepthAnything (finetuned)} & \textit{0.107} & \textit{0.831} & \textit{3.866} & \textit{0.185} & \textit{0.890} & \textit{0.950} & \textit{0.974} & - & \textit{335.79} M & \textit{130.41} ms \\
        \midrule
        \multirow{6}{*}{\rotatebox[origin=c]{90}{Self-supervised}} & Monodepth2~\cite{godard2019digging} & 0.287 & 3.349 & 7.184 & 0.345 & 0.641 & 0.845 & 0.925 & - & 14.33 M & 3.44 ms \\
        & PackNet-SfM~\cite{guizilini20203d} & 0.309 & 2.891 & 7.994 & 0.390 & 0.547 & 0.796 & 0.899 & - & 128.29 M & 57.86 ms \\
        & FSM~\cite{guizilini2022full} & 0.299 & - & - & - & - & - & - & - & 14.33 M & 3.44 ms \\
        & SurroundDepth~\cite{wei2023surrounddepth} & 0.245 & 3.067 & 6.835 & 0.321 & 0.719 & 0.878 & 0.935 & - & 59.56 M & 10.82 ms\\
        & EGA-Depth-LR~\cite{shi2023ega} & 0.239 & 2.357 & 6.801 & 0.317 & 0.723 & 0.880 & 0.936 & - & not public & not public \\
        & EGA-Depth-MR~\cite{shi2023ega} & \textbf{0.228} & \textbf{2.113} & \textbf{6.738} & \textbf{0.311} & \textbf{0.728} & \textbf{0.885} & \textbf{0.940} & - & not public & not public \\
        \midrule
        \multirow{9}{*}{\rotatebox[origin=c]{90}{Supervised}} & Monodepth2 & 0.182 & 1.593 & 5.961 & 0.295 & 0.744 & 0.868 & 0.926 & 0.00 & 14.33 M & 3.44 ms \\
        & Monodepth2 + MB & 0.188 & 1.719 & 6.145 & 0.311 & 0.733 & 0.857 & 0.917 & $-$3.35 & 15.03 M & 11.71 ms \\
        & Monodepth2 + MB + Ours & \textbf{0.158} & \textbf{1.135} & \textbf{4.915} & \textbf{0.243} & \textbf{0.805} & \textbf{0.914} & \textbf{0.957} & \textbf{$+$13.42} & 15.03 M & 11.71 ms \\
        \cmidrule{2-12}
        & HRDepth & 0.172 & 1.313 & 5.523 & 0.267 & 0.765 & 0.893 & 0.946 & 0.00 & 14.10 M & 5.67 ms \\
        & HRDepth + MB & 0.174 & 1.360 & 5.439 & 0.278 & 0.762 & 0.881 & 0.934 & $-$1.48 & 14.80 M & 13.92 ms \\
        & HRDepth + MB + Ours & \textbf{0.156} & \textbf{1.097} & \textbf{5.020} & \textbf{0.244} & \textbf{0.792} & \textbf{0.911} & \textbf{0.958} & \textbf{$+$7.18} & 14.80 M & 13.92 ms \\
        % \midrule
        \cmidrule{2-12}
        & MonoViT & 0.170 & 1.643 & 5.174 & 0.253 & 0.769 & 0.884 & 0.940 & 0.00 & 27.87 M & 17.26 ms \\
        & MonoViT + MB & 0.169 & 1.616 & 5.121 & 0.252 & 0.770 & 0.884 & 0.940 & $+$0.54 & 28.31 M & 25.40 ms \\
        & MonoViT + MB + Ours & \textbf{0.134} & \textbf{1.072} & \textbf{4.096} & \textbf{0.214} & \textbf{0.830} & \textbf{0.914} & \textbf{0.954} & \textbf{$+$15.00} & 28.31 M & 25.40 ms \\
        % \midrule
        % Adabins & & & & & & & \\
        % Adabins + CKD & & & & & & & \\
        % \midrule
        % binsformer & & & & & & & \\
        % binsformer + CKD & & & & & & & \\
        % \midrule
        % MonoViT & & & & & & & \\
        % MonoViT + LocalBins & & & & & & & \\
        % MonoViT + LocalBins + CKD & & & & & & & \\
        \bottomrule
    \end{tabular}}
\end{table*}

% \begin{table*}[t]
%     \caption{Comparision between existing KD approaches and our method on \textbf{DDAD} dataset. All experiments are conducted on Monodepth2 architectures. $\Delta_\mathcal{T}$ implies the relative performance increment from Monodepth2.}
%     \label{tab:kd}
%     \centering
%     \resizebox{0.76\linewidth}{!}{
%     \begin{tabular}{lcccccccr}
%         \toprule
%         Methods & Abs Rel & Sq Rel & RMSE & RMSE log & $\delta < 1.25$ & $\delta < 1.25^2$ & $\delta < 1.25^3$ & $\Delta_\mathcal{T} (\%)$ \\
%         % Methods & Abs Rel $\downarrow$ & Sq Rel $\downarrow$ & RMSE $\downarrow$ & RMSE log $\downarrow$ & $\delta < 1.25 \uparrow$ & $\delta < 1.25^2 \uparrow$ & $\delta < 1.25^3 \uparrow$ & $\Delta_\mathcal{T} (\%) \uparrow$ \\
%         \midrule
%         Monodepth2 & 0.200 & 3.087 & 12.849 & 0.323 & 0.679 & 0.861 & 0.932 & 0.00 \\
%         \midrule
%         KD~\cite{hinton2015distilling} & 0.222 & 3.444 & 13.017 & 0.332 & 0.662 & 0.852 & 0.928 & $-$4.38 \\
%         FitNets~\cite{romero2014fitnets} & 0.200 & 2.993 & 12.623 & 0.308 & 0.688 & 0.872 & 0.940 & $+$1.84 \\
%         AT~\cite{zagoruyko2016paying} & 0.195 & 2.997 & 12.505 & 0.311 & 0.694 & 0.875 & 0.935 & $+$2.28  \\
%         SP~\cite{tung2019similarity} & 0.194 & 2.931 & 12.331 & 0.308 & 0.692 & 0.876 & 0.941 & $+$3.05 \\
%         \midrule
%         Ours & \textbf{0.191} & \textbf{2.865} & \textbf{12.134} & \textbf{0.300} & \textbf{0.710} & \textbf{0.881} & \textbf{0.943} & \textbf{$+$4.64} \\
%         \bottomrule
%     \end{tabular}}
% \end{table*}

\begin{table*}[t]
    \caption{Comparision between existing KD approaches and our method on \textbf{DDAD} dataset. All experiments are conducted on Monodepth2 architectures. $\Delta_\mathcal{T}$ implies the relative performance increment from Monodepth2.}
    \label{tab:kd}
    \centering
    \resizebox{0.76\linewidth}{!}{
    \begin{tabular}{lcccccccr}
        \toprule
        % Methods & Abs Rel \textcolor{red}{$\downarrow$} & Sq Rel \textcolor{red}{$\downarrow$} & RMSE \textcolor{red}{$\downarrow$} & RMSE log \textcolor{red}{$\downarrow$} & $\delta < 1.25$ \textcolor{red}{$\uparrow$} & $\delta < 1.25^2$ \textcolor{red}{$\uparrow$} & $\delta < 1.25^3$ \textcolor{red}{$\uparrow$} & $\Delta_\mathcal{T} (\%)$ \textcolor{red}{$\uparrow$} \\
        Methods & Abs Rel $\downarrow$ & Sq Rel $\downarrow$ & RMSE $\downarrow$ & RMSE log $\downarrow$ & $\delta < 1.25$ $\uparrow$ & $\delta < 1.25^2$ $\uparrow$ & $\delta < 1.25^3$ $\uparrow$ & $\Delta_\mathcal{T} (\%)$ $\uparrow$ \\
        % Methods & Abs Rel $\downarrow$ & Sq Rel $\downarrow$ & RMSE $\downarrow$ & RMSE log $\downarrow$ & $\delta < 1.25 \uparrow$ & $\delta < 1.25^2 \uparrow$ & $\delta < 1.25^3 \uparrow$ & $\Delta_\mathcal{T} (\%) \uparrow$ \\
        \midrule
        Monodepth2 & 0.200 & 3.087 & 12.849 & 0.323 & 0.679 & 0.861 & 0.932 & 0.00 \\
        \midrule
        KD~\cite{hinton2015distilling} & 0.222 & 3.444 & 13.017 & 0.332 & 0.662 & 0.852 & 0.928 & $-$4.38 \\
        FitNets~\cite{romero2014fitnets} & 0.200 & 2.993 & 12.623 & 0.308 & 0.688 & 0.872 & 0.940 & $+$1.84 \\
        AT~\cite{zagoruyko2016paying} & 0.195 & 2.997 & 12.505 & 0.311 & 0.694 & 0.875 & 0.935 & $+$2.28  \\
        SP~\cite{tung2019similarity} & 0.194 & 2.931 & 12.331 & 0.308 & 0.692 & 0.876 & 0.941 & $+$3.05 \\
        \midrule
        Ours & \textbf{0.191} & \textbf{2.865} & \textbf{12.134} & \textbf{0.300} & \textbf{0.710} & \textbf{0.881} & \textbf{0.943} & \textbf{$+$4.64} \\
        \bottomrule
    \end{tabular}}
    % \vspace{-8pt}
\end{table*}

\subsection{Evaluation Results on FSMDE Datasets}
\tabref{tab:ddad} summarizes the quantitative evaluation results on the DDAD dataset.
We evaluate each baseline (\textit{i.e.,} Monodepth2, HRDepth, and MonoViT) by applying the MetricBins module and our proposed method.
Given that the majority of FSMDE research employs self-supervised methods, we additionally reported the performance of self-supervised FSMDE approaches in our table for a general comparison.
The results indicate that even with post-processing (i.e., median scaling), self-supervised methodologies generally exhibit lower performance compared to those trained in a supervised manner.
% 대부분의 FSMDE 연구가 self-supervised method를 쓰기 때문에 대략적인 비교를 위해 우리는 self-supervised FSMDE의 성능을 테이블에 추가적으로 report했다.
% 그 결과 self-supervised 방법론들은 post-processing (i.e., median scaling)을 거쳤음에도 불구하고 supervised manner로 학습된 방법론보다 대체로 낮은 성능을 보였다.
When the MetricBins method is added to each baseline, the average improvement ranges from as low as 0.28\% to as high as 1.08\% compared to the pure baseline.
Compared to the improvements achieved by MetricBins, applying the proposed methods (\textit{i.e.}, $\mathcal{L}_{\text{ckd}}$ and $\mathcal{L}_{\text{vrkd}}$) leads to markedly greater improvements, an average improvement ranging from 4.64\% to 7.54\%.
% Furthermore, as shown in \figref{fig:qualitative_ddad}, our method, which transfers probability-level knowledge from the foundation model serving as the teacher network, demonstrates enhanced robustness in both long-range and fine detail compared to the baseline.
Furthermore, as illustrated in \figref{fig:qualitative_ddad}, transferring probability-level knowledge from the teacher network contributes to improved generalization performance in regions lacking LiDAR points, which are problematic for supervised MDE.
% Furthermore, as shown in \figref{fig:qualitative_ddad}, by transferring the probability-level knowledge from the teacher network, the lidar point가 없는 부분 (e.g., sky)에 대한 generalization performance에 도움이 된다.

On the other hand, applying the MetricBins to Monodepth2 and HRDepth results in slight performance degradation on the nuScenes dataset, as shown in \tabref{tab:nuscenes}.
Although the performance drop is minor, this indicates that simply incorporating the binning strategy alone does not lead to significant improvements.
In contrast, the substantial performance gain achieved by our method, which distills knowledge at the bin probability level, further validates its effectiveness.
The performance improvement ranges from 7.18\% to 15\%, with the largest gain observed in MonoViT, while our method also satisfies real-time latency requirements (\textit{i.e.,} under 33ms), highlighting its practical applicability.
These results demonstrate our method's effectiveness and robustness across different datasets and model architectures.

\begin{table*}[t]
    \caption{Evaluation results of various binning methods and our method on \textbf{DDAD} dataset. All experiments are conducted on Monodepth2 architectures. (AB: AdaBins, LB: LocalBins, MB: MetricBins)}
    \centering
    \resizebox{1.00\linewidth}{!}{
    \label{tab:bins}
    \begin{tabular}{lcccccccr|rr}
        \toprule
        % Methods & Abs Rel \textcolor{red}{$\downarrow$} & Sq Rel \textcolor{red}{$\downarrow$} & RMSE \textcolor{red}{$\downarrow$} & RMSE log \textcolor{red}{$\downarrow$} & $\delta < 1.25$ \textcolor{red}{$\uparrow$} & $\delta < 1.25^2$ \textcolor{red}{$\uparrow$} & $\delta < 1.25^3$ \textcolor{red}{$\uparrow$} & $\Delta_\mathcal{T} (\%)$ \textcolor{red}{$\uparrow$} & Params. $\downarrow$ & Latency $\downarrow$ \\
        Methods & Abs Rel $\downarrow$ & Sq Rel $\downarrow$ & RMSE $\downarrow$ & RMSE log $\downarrow$ & $\delta < 1.25$ $\uparrow$ & $\delta < 1.25^2$ $\uparrow$ & $\delta < 1.25^3$ $\uparrow$ & $\Delta_\mathcal{T} (\%)$ $\uparrow$ & Params. $\downarrow$ & Latency $\downarrow$ \\
        \midrule
        Monodepth2 & 0.200 & 3.087 & 12.849 & 0.323 & 0.679 & 0.861 & 0.932 & 0.00 & 14.33 M & 3.75 ms \\
        \midrule
        Monodepth2 + AB & \textbf{0.502} & \textbf{5.004} & \textbf{13.421} & \textbf{0.471} & \textbf{0.377} & \textbf{0.659} & \textbf{0.828} & \textbf{$-$48.92} & 16.36 M & 5.46 ms \\
        Monodepth2 + AB + Ours & 1.194 & 14.074 & 16.496 & 0.789 & 0.167 & 0.342 & 0.531 & $-$172.04 & 16.36 M & 5.46 ms \\
        \midrule
        Monodepth2 + LB & 0.194 & 3.011 & 12.880 & 0.317 & 0.699 & 0.871 & 0.933 & $+$1.61 & 18.20 M & 10.58 ms \\
        Monodepth2 + LB + Ours & \textbf{0.191}  & \textbf{2.819} & \textbf{12.446} & \textbf{0.295} & \textbf{0.716} & \textbf{0.884} & \textbf{0.945} & \textbf{$+$4.93} & 18.20 M & 10.58 ms \\
        \midrule
        Monodepth2 + MB & 0.195 & 3.016 & 12.727 & 0.322 & 0.686 & 0.865 & 0.932 & $+$1.08 & 15.03 M & 12.77 ms \\
        Monodepth2 + MB + Ours & \textbf{0.191} & \textbf{2.865} & \textbf{12.134} & \textbf{0.300} & \textbf{0.710} & \textbf{0.881} & \textbf{0.943} & \textbf{$+$4.64} & 15.03 M & 12.77 ms \\
        % Monodepth + MB & & & & & & & & \\
        % Monodepth + MB + Ours & & & & & & & & \\
        \bottomrule
    \end{tabular}}
    % \vspace{-5pt}
\end{table*}

\subsection{Comparison between Our Method and Existing Knowledge Distillation Methods}
\label{subsec:comparison_kd}

We compare the existing knowledge distillation approaches with our proposed method to demonstrate our probability-level knowledge distillation method.
% The knowledge distillation baselines consist of conventional output-level KD~\cite{hinton2015distilling} and three feature-level knowledge distillation methods, including FitNets~\cite{romero2014fitnets}, AT~\cite{zagoruyko2016paying}, and SP~\cite{tung2019similarity}.
Due to that, recent distillation methods often leverage class priors or inter-data relationships~\cite{tian2019contrastive, chen2022knowledge}---particularly function for only classification tasks---we incorporate general distillation methods: conventional output-level KD~\cite{hinton2015distilling} and three feature-level distillation baselines, including FitNets~\cite{romero2014fitnets}, AT~\cite{zagoruyko2016paying}, and SP~\cite{tung2019similarity}.
In the current experimental setup, where accurate metric scale depth estimation is required, a scale-ambiguous teacher model can hinder a student network from learning metric depth properly.
To prevent this issue and for a proper comparison, we do not include output-level distillation for all distillation approaches except for KD, which inherently operates at the output level.
Furthermore, considering the low transferability in cross-architecture settings (\textit{i.e.}, a transformer teacher model and a CNN student model~\cite{liu2022cross}), we apply feature-level distillation only between the CNN decoders of the teacher and student models. 

\tabref{tab:kd} summarizes the evaluation results of the proposed method along with other knowledge distillation approaches.
Due to scale ambiguity in the output-level of the teacher foundation model (refer to \secref{sec:intro}), KD severely degrades performance compared to a fine-tuned student network.
Our proposed method, which performs distillation at the scale-invariant bin probability level, achieves the best performance while other feature-level distillation methods show slight improvements.
These results further validate the effectiveness of our approach in mitigating scale ambiguity while maximizing the benefits of knowledge distillation.

\subsection{Evaluation Results of Various Binning Methods}
\label{subsec:comparion_bin}

To assess the effectiveness of the proposed method in relation to different binning strategies, we integrate various binning methods, including AdaBins~\cite{bhat2021adabins}, LocalBins~\cite{bhat2022localbins}, and MetricBins~\cite{bhat2023zoedepth}, into our strategy for comparison.
Specifically, AdaBins leverages global adaptive depth bin centers $C_{\text{AdaBins}} \in \mathbb{R}^{B}$ that are shared across all pixels, as opposed to per-pixel adaptive depth bin centers $C_{\text{LocalBins}}, C_{\text{MetricBins}} \in \mathbb{R}^{B \times H \times W}$, where $B$ denotes the number of depth bins.
As detailed in \equref{eq:depth_recon}, each binning module reconstructs the depth map by utilizing the depth bin probability $P \in \mathbb{R}^{B \times H \times W}$ along with the corresponding global or local depth bin centers.
Due to these methodological differences, AdaBins produces a depth bin probability that reflects the overall depth structure of the scene, whereas LocalBins and MetricBins yield locally adaptive probability distributions.

\tabref{tab:bins} presents the results of applying each binning method and our method to Monodepth2.
The use of AdaBins leads to significant performance degradation because the global binning strategy of AdaBins does not align with the teacher model's locally adaptive nature, leading to improper knowledge distillation of the probability distribution.
% The use of AdaBins leads to significant performance degradation, as during distillation, the bin configuration between the teacher and student models is aligned by reducing the bin count estimated by AdaBins, which limits the granularity required to partition a wide depth range.
% Furthermore, the global binning strategy of AdaBins does not align with the teacher model's locally adaptive nature, leading to improper knowledge distillation of the probability distribution.
Meanwhile, LocalBins and MetricBins, which adopt the same pixel-level mechanism as the teacher model, enable effective knowledge distillation, achieving relative improvements of 4.93\% and 4.64\%, respectively.

\subsection{Ablation Studies of Our Method}
\label{subsec:ablation}
We conduct ablation studies on two key components of our method: $\mathcal{L}_{\text{ckd}}$ and $\mathcal{L}_{\text{vrkd}}$, which play a crucial role in distilling knowledge at the bin probability level.
To investigate the impact of these losses, we evaluate the performance of Monodepth2, HRDepth, and MonoViT on the DDAD dataset by comparing results with and without each loss term.
% As shown in \tabref{tab:ablation}, employing MB alone yields a 1.08\% improvement over Monodepth2, with improvements of 0.28\% for HRDepth and 0.34\% for MonoViT.
As shown in \tabref{tab:ablation}, employing MB alone yields an average improvement of 0.57\% over all models.
% 1.08\% improvement over Monodepth2, with improvements of 0.28\% for HRDepth and 0.34\% for MonoViT.
Integrating $\mathcal{L}_{\text{ckd}}$ with MB further improves the performance, achieving a total gain of 3.92\%, 5.12\%, and 6.37\% for Monodepth2, HRDepth, and MonoViT, respectively, which highlights the effectiveness of transferring the teacher foundation model’s bin probability distribution to the student model.

Furthermore, incorporating $\mathcal{L}_{\text{vrkd}}$ results in an additional enhancement, leading to an overall 4.64\%, 5.47\%, and 7.54\% performance gain for each model, and these consistent performance improvements demonstrate the stability and robustness of our proposed method.
This result suggests that View-relational Knowledge Distillation (\secref{subsec:viewkd}) is complementary to Cross-interaction Knowledge Distillation (\secref{subsec:ckd}), further enhancing depth estimation performance by leveraging view relational constraints.% The complete evaluation tables are provided in the supplementary material.
% \begin{minipage}
\begin{table}[t]
    \caption{Ablation studies of our methods on \textbf{DDAD} dataset.}
    \centering
    \resizebox{0.9\linewidth}{!}{
    \label{tab:ablation}
    \begin{tabular}{cccrrr}
        \toprule
        % \multicolumn{3}{c}{Methods} & $\Delta_\mathcal{T} (\%) \uparrow$ \\
        % \midrule
        \multirow{2}{*}{MB} & \multirow{2}{*}{$\mathcal{L}_{\text{ckd}}$} & \multirow{2}{*}{$\mathcal{L}_{\text{vrkd}}$} & Monodepth2 & HRDepth & MonoViT \\
        \cmidrule{4-6}
        & & & $\Delta_\mathcal{T} (\%) \uparrow$ & $\Delta_\mathcal{T} (\%) \uparrow$ & $\Delta_\mathcal{T} (\%) \uparrow$\\
        % \cmidrule{2-5}
        \midrule
        \ding{55} & \ding{55} & \ding{55} & 0.00 & 0.00 & 0.00\\
        \ding{51} & \ding{55} & \ding{55} & $+$1.08 & $+$0.28 & $+$0.34\\
        \ding{51} & \ding{51} & \ding{55} & $+$3.92 & $+$5.12 & $+$6.37 \\
        % \textcolor{red}{\ding{51}} & \textcolor{red}{\ding{55}} & \textcolor{red}{\ding{51}} & \textcolor{red}{$+$0.83} & \textcolor{red}{$+$0.83} & \textcolor{red}{$+$1.11} \\
        \ding{51} & \ding{55} & \ding{51} & $+$0.83 & $+$0.83 & $+$1.11 \\
        \ding{51} & \ding{51} & \ding{51} & \textbf{$+$4.64} & \textbf{$+$5.47} & \textbf{$+$7.54} \\
        % \midrule
        % \multirow{4}{*}{\rotatebox[origin=c]{90}{HRDepth}} & & & & 0.00\\
        % & \checkmark & & & $+$0.28\\
        % & \checkmark & \checkmark & & \textbf{$+$5.90} \\
        % & \checkmark & \checkmark & \checkmark & $+$5.47 \\
        % \midrule
        % \multirow{4}{*}{\rotatebox[origin=c]{90}{MonoViT}} & & & & 0.00\\
        % & \checkmark & & & $+$0.34\\
        % & \checkmark & \checkmark & & $+$6.37\\
        % & \checkmark & \checkmark & \checkmark & \textbf{$+$7.54} \\
        \bottomrule
    \end{tabular}}
    % \vspace{-8pt}
\end{table}
\section{Conclusion}
\label{sec:conclusion}
In this paper, we propose a new knowledge distillation strategy for the FSMDE framework that transfers a robust depth representation from a foundation model to a lightweight FSMDE student model suitable for autonomous driving scenarios.
Our method is based on the predominant binning strategy used in supervised MDE methodologies, and can be applied seamlessly.
The proposed cross-interaction knowledge distillation can effectively transfer the well-structured relative depth knowledge of the teacher foundation model to the student model, which enables the student model to mimic the teacher's robust representation.
View-relational Knowledge Distillation enables the student model to learn the relational knowledge of adjacent camera views by injecting the teacher's structural probabilities into the student model.
Experiments on DDAD and nuScenes demonstrate that our method achieves significant performance improvements while maintaining real-time processing at 30 FPS.
The results also demonstrate superior performance compared to existing distillation methods, highlighting the effectiveness of our approach in combining a binning strategy with FSMDE.
% \section{Conclusion}
% We have presented TaskForce, a cooperative MARL approach that reformulates multi-task learning as a multi-objective optimization problem.
% By assigning each task to an agent, which observes compact gradient information and loss convergence signals and receives a loss-gradient hybrid reward, TaskForce effectively learns to balance task tradeoffs and resolve gradient conflicts.
% Empirical results on NYU-v2, Cityscapes, and QM9 demonstrate that TaskForce substantially outperforms competing baselines, yielding more stable and effective multi-task training.
% This work highlights the promise of integrating reinforcement learning with gradient-based optimization for MTL.
% Future extensions of TaskForce may involve scaling to even larger task sets and exploring richer reward structures, further validating the potential of cooperative agent-based optimization in deep learning.

\bibliographystyle{IEEEtran}
\bibliography{IEEEexample}
\clearpage
\onecolumn
% \appendix

\noindent\textbf{\LARGE APPENDIX}
% \input{supp_tables/table_ssfsmde}

% \section{Comparison Between Self-Supervised FSMDE Methods and the Proposed Method}  

% Since the introduction of FSM~\cite{guizilini2022full}, the majority of research efforts in Full Surround Monocular Depth Estimation (FSMDE) have predominantly focused on self-supervised learning frameworks~\cite{kim2022self, wei2023surrounddepth, shi2023ega, schmied2023r3d3}.
% These approaches leverage geometric consistency across multiple views and spatio-temporal cues to refine depth predictions without requiring explicit ground-truth supervision.
% In contrast, supervised FSMDE approaches~\cite{guo2023simple} have been relatively underexplored, primarily due to the high cost of obtaining dense and accurate depth annotations.
% Given this research gap, we present a comparative evaluation of our proposed method against self-supervised FSMDE approaches in \tabref{tab:ssfsmde_ddad}-\ref{tab:ssfsmde_nuscene}.

\begin{table*}[h]
    \caption{Evaluation results of our method with respect to camera views on the \textbf{nuScenes} dataset, where all metrics are Abs Rel and MB refers to MetricBins~\cite{bhat2023zoedepth}.}
    \label{tab:viewwise_nuscenes}
    \centering
    \resizebox{0.8\linewidth}{!}{
    \begin{tabular}{lcccccccr}
        \toprule
        Methods & Front & F. Left & F. Right & B. Left & B. Right & Back & Total \\
        \midrule
        Monodepth2~\cite{godard2019digging} & 0.131 & 0.185 & 0.228 & 0.197 & 0.233 & 0.115 & 0.182 \\
        Monodepth2 + MB & 0.145 & 0.182 & 0.240 & 0.196 & 0.246 & 0.121 & 0.188\\
        Monodepth2 + MB + Ours & \textbf{0.103} & \textbf{0.180} & \textbf{0.190} & \textbf{0.186} & \textbf{0.202} & \textbf{0.084} & \textbf{0.158} \\
        \midrule
        HRDepth~\cite{lyu2021hr} & 0.118 & 0.183 & 0.204 & 0.193 & 0.208 & 0.129 & 0.172 \\
        HRDepth + MB & 0.123 & \textbf{0.178} & 0.208 & 0.189 & 0.220 & 0.129 & 0.174 \\
        HRDepth + MB + Ours & \textbf{0.104} & 0.180 & \textbf{0.191} & \textbf{0.184} & \textbf{0.193} & \textbf{0.084} & \textbf{0.156} \\
        \midrule
        MonoViT~\cite{zhao2022monovit} & 0.103 & 0.188 & 0.140 & 0.190 & 0.196 & 0.204 & 0.170 \\
        MonoViT + MB & 0.101 & 0.188 & 0.138 & 0.191 & 0.195 & 0.202 & 0.169 \\
        MonoViT + MB + Ours & \textbf{0.077} & \textbf{0.153} & \textbf{0.134} & \textbf{0.174} & \textbf{0.184} & \textbf{0.081} & \textbf{0.134} \\
        \bottomrule
    \end{tabular}}
\end{table*}

\section{Performance Analysis of Our Method Across Different Camera Views}

To further demonstrate the effectiveness of our approach, we conduct a detailed performance analysis with respect to individual camera views in the nuScenes dataset.
A comprehensive summary of the evaluation results is presented in \tabref{tab:viewwise_nuscenes}, where we report Absolute Relative Error (Abs Rel) for each camera.
Similar to the trends in the experiments in our main manuscript, our method consistently enhances depth estimation performance across all baseline models and camera views.
These results reinforce the versatility of our method, highlighting its ability to consistently improve monocular depth estimation across a full surround camera system in real-world autonomous driving scenarios.
% Similar to the trends observed in our main experiments, our method consistently enhances depth estimation accuracy across all baseline models and camera viewpoints.
% Notably, the performance gains are particularly pronounced in cameras with wider fields of view, suggesting that our knowledge distillation framework effectively transfers spatial information across multiple perspectives.

\begin{table*}[!ht]
    \caption{Evaluation results of DepthAnything~\cite{yang2024depth} under the various configurations on the \textbf{DDAD} dataset. (Z.S.:Zeroshot, F.T.:Fine-Tuned)}
    \label{tab:depthany_ddad}
    \centering
    \resizebox{1.00\linewidth}{!}{
    \begin{tabular}{llccccccc}
        \toprule
        \multicolumn{2}{l}{Methods} & Abs Rel $\downarrow$ & Sq Rel $\downarrow$ & RMSE $\downarrow$ & RMSE log $\downarrow$ & $\delta < 1.25 \uparrow$ & $\delta < 1.25^2 \uparrow$ & $\delta < 1.25^3 \uparrow$ \\
        \midrule
        \multirow{3}{*}{\rotatebox[origin=c]{90}{Z.S.}} & DepthAnything-S & 0.287 & 3.571 & 12.050 & 0.634 & 0.577 & 0.810 & 0.898 \\
        & DepthAnything-B & 0.281 & 3.412 & 12.016 & 0.652 & 0.580 & 0.816 & 0.898 \\
        & DepthAnything-L & \textbf{0.270} & \textbf{3.291} & \textbf{11.866} & \textbf{0.621} & \textbf{0.601} & \textbf{0.826} & \textbf{0.904} \\
        & UniDepth~\cite{piccinelli2024unidepth} & 0.172 & 2.057 & 10.100 & 0.247 & 0.804 & 0.931 & 0.967 \\
        \midrule
        \multirow{3}{*}{\rotatebox[origin=c]{90}{F.T.}} & DepthAnything-S & 0.160 & 2.178 & 10.284 & 0.248 & 0.792 & 0.923 & 0.963 \\
        & DepthAnything-B & 0.148 & 1.995 & 9.860 & 0.237 & 0.815 & 0.930 & 0.967 \\
        & DepthAnything-L & \textbf{0.140} & \textbf{1.866} & \textbf{9.475} & \textbf{0.228} & \textbf{0.831} & \textbf{0.935} & \textbf{0.969} \\
        & UniDepth~\cite{piccinelli2024unidepth} & 0.162 & 1.990 & 9.880 & 0.242 & 0.804 & 0.928 & 0.966 \\
        \bottomrule
    \end{tabular}}
    \caption{Evaluation results of DepthAnything~\cite{yang2024depth} under the various configurations on the \textbf{nuScenes} dataset. (Z.S.:Zeroshot, F.T.:Fine-Tuned)}
    \label{tab:depthany_nuscenes}
    \vspace{5pt}
    \resizebox{1.00\linewidth}{!}{
    \begin{tabular}{llccccccc}
        \toprule
        \multicolumn{2}{l}{Methods} & Abs Rel $\downarrow$ & Sq Rel $\downarrow$ & RMSE $\downarrow$ & RMSE log $\downarrow$ & $\delta < 1.25 \uparrow$ & $\delta < 1.25^2 \uparrow$ & $\delta < 1.25^3 \uparrow$ \\
        \midrule
        \multirow{3}{*}{\rotatebox[origin=c]{90}{Z.S.}} & DepthAnything-S & 0.232 & 1.537 & 5.482 & 0.301 & 0.690 & 0.882 & 0.944 \\
        & DepthAnything-B & 0.214 & 1.372 & 5.221 & 0.284 & 0.723 & 0.896 & 0.950 \\
        & DepthAnything-L & \textbf{0.208} & \textbf{1.324} & \textbf{5.108} & \textbf{0.280} & \textbf{0.735} & \textbf{0.901} & \textbf{0.951} \\
        & UniDepth~\cite{piccinelli2024unidepth} & 0.158 & 1.066 & 4.689 & 0.238 & 0.809 & 0.921 & 0.963 \\
        \midrule
        \multirow{3}{*}{\rotatebox[origin=c]{90}{F.T.}} & DepthAnything-S & 0.134 & 0.926 & 4.544 & 0.213 & 0.844 & 0.937 & 0.970 \\
        & DepthAnything-B & 0.115 & 0.796 & 4.071 & 0.191 & 0.879 & \textbf{0.952} & \textbf{0.975} \\
        & DepthAnything-L & \textbf{0.107} & \textbf{0.831} & \textbf{3.866} & \textbf{0.185} & \textbf{0.890} & 0.950 & 0.974 \\
        & UniDepth~\cite{piccinelli2024unidepth} & 0.121 & 0.791 & 3.833 & 0.190 & 0.879 & 0.951 & 0.976 \\
        \bottomrule
    \end{tabular}}
\end{table*}

\section{Full Evaluation Results of MDE Foundation Models}

In our main manuscripts, we leverage the DepthAnything~\cite{yang2024depth}, particularly, pre-trained DepthAnything-L model as a teacher model for all experiments.
% In \tabref{tab:depthany_ddad}-\ref{tab:depthany_nuscenes}, we report the full evaluation results of DepthAnything under the different configurations of model parameters.
To provide a more comprehensive evaluation, we report the full performance results of DepthAnything with various configurations of model parameters, and additional MDE foundation model, UniDepth~\cite{piccinelli2024unidepth} in \tabref{tab:depthany_ddad}-\ref{tab:depthany_nuscenes}.
We apply the affine-invariant depth normalization~\cite{ke2024repurposing, yang2024depth} for zero-shot evaluation of DepthAnything, and do not apply any scaling for zero-shot UniDepth and all fine-tuned models, respectively.
% Note that these foundation models pretrained by training on a combination of 1.5M labeled images and 62M$+$ unlabeled images for robust monocular depth estimation.
It is important to note that these foundation models have been pre-trained on a large-scale dataset, comprising a combination of over 1.5 million labeled images and more than 62 million unlabeled images.

\begin{table*}[h]
    \caption{Ablation studies of the proposed methods on \textbf{DDAD} dataset with full metrics. $\Delta_\mathcal{T}$ denotes the relative performance improvement over baseline models.}
    \label{tab:ablation_supp}
    \centering
    \resizebox{1.0\linewidth}{!}{
    \begin{tabular}{ccccrrrrrrrr}
        \toprule
        % \multicolumn{3}{c}{Methods} & $\Delta_\mathcal{T} (\%) \uparrow$ \\
        % \midrule
        & MB & $\mathcal{L}_{\text{ckd}}$ & $\mathcal{L}_{\text{vrkd}}$ & Abs Rel $\downarrow$ & Sq Rel $\downarrow$ & RMSE $\downarrow$ & RMSE log $\downarrow$ & $\delta < 1.25 \uparrow$ & $\delta < 1.25^2 \uparrow$ & $\delta < 1.25^3 \uparrow$ & $\Delta_{\mathcal{T}}(\%) \uparrow$ \\
        \midrule
        \multirow{4}{*}{\rotatebox[origin=c]{90}{\small Monodepth2}} & \ding{55} & \ding{55} & \ding{55} & 0.200 & 3.087 & 12.849 & 0.323 & 0.679 & 0.861 & 0.932 & 0.00 \\
        & \ding{51} & \ding{55} & \ding{55} & 0.195 & 3.016 & 12.727 & 0.322 & 0.686 & 0.865 & 0.932 & $+$1.08 \\
        & \ding{51} & \ding{51} & \ding{55} & 0.193 & 2.899 & 12.238 & 0.305 & 0.708 & 0.880 & 0.942 & $+$3.92 \\
        & \ding{51} & \ding{51} & \ding{51} & \textbf{0.191} & \textbf{2.865} & \textbf{12.134} & \textbf{0.300} & \textbf{0.710} & \textbf{0.881} & \textbf{0.943} & \textbf{$+$4.64} \\
        \midrule
        \multirow{4}{*}{\rotatebox[origin=c]{90}{HRDepth}} & \ding{55} & \ding{55} & \ding{55} & 0.196 & 2.955 & 12.428 & 0.304 & 0.699 & 0.875 & 0.940 & 0.00 \\
        & \ding{51} & \ding{55} & \ding{55} & 0.194 & 2.961 & 12.399 & 0.303 & 0.700 & 0.877 & 0.942 & $+$0.28 \\
        & \ding{51} & \ding{51} & \ding{55} & 0.184 & \textbf{2.681} & 11.580 & 0.285 & 0.731 & 0.893 & 0.947 & $+$5.12 \\
        & \ding{51} & \ding{51} & \ding{51} & \textbf{0.183} & 2.684 & \textbf{11.578} & \textbf{0.283} & \textbf{0.736} & \textbf{0.896} & \textbf{0.950} & \textbf{$+$5.47} \\
        \midrule
        \multirow{4}{*}{\rotatebox[origin=c]{90}{MonoViT}} & \ding{55} & \ding{55} & \ding{55} & 0.179 & 2.602 & 11.777 & 0.287 & 0.725 & 0.892 & 0.949 & 0.00 \\
        & \ding{51} & \ding{55} & \ding{55} & 0.178 & 2.585 & 11.757 & 0.286 & 0.728 & 0.893 & 0.950 & $+$0.34 \\
        & \ding{51} & \ding{51} & \ding{55} & 0.168 & 2.320 & 10.720 & 0.260 & 0.768 & 0.912 & 0.959 & $+$6.37 \\
        & \ding{51} & \ding{51} & \ding{51} & \textbf{0.166} & \textbf{2.260} & \textbf{10.483} & \textbf{0.256} & \textbf{0.773} & \textbf{0.916} & \textbf{0.961} & \textbf{$+$7.54} \\
        \bottomrule
    \end{tabular}}
\end{table*}

\section{Full Evaluation Results of Ablation Studies}
We summarize the full evaluation results of ablation studies, which are discussed in \secref{subsec:ablation}, in \tabref{tab:ablation_supp}.
% The experiment incorporating all key components of our method demonstrates its effectiveness by achieving the best performance in 20 out of 21 metrics across all baselines.
The experimental results clearly demonstrate that incorporating all key components of our method yields the most significant performance gains.
Specifically, our approach achieves the best results in 20 out of 21 evaluation metrics across all baselines, confirming the effectiveness of our knowledge distillation strategy in enhancing monocular depth estimation for full surround camera systems.

\section{Qualitative Results of Our Method on DDAD and nuScenes Datasets}
\begin{figure*}[!t]
    \centering
    \includegraphics[width=0.82\linewidth]{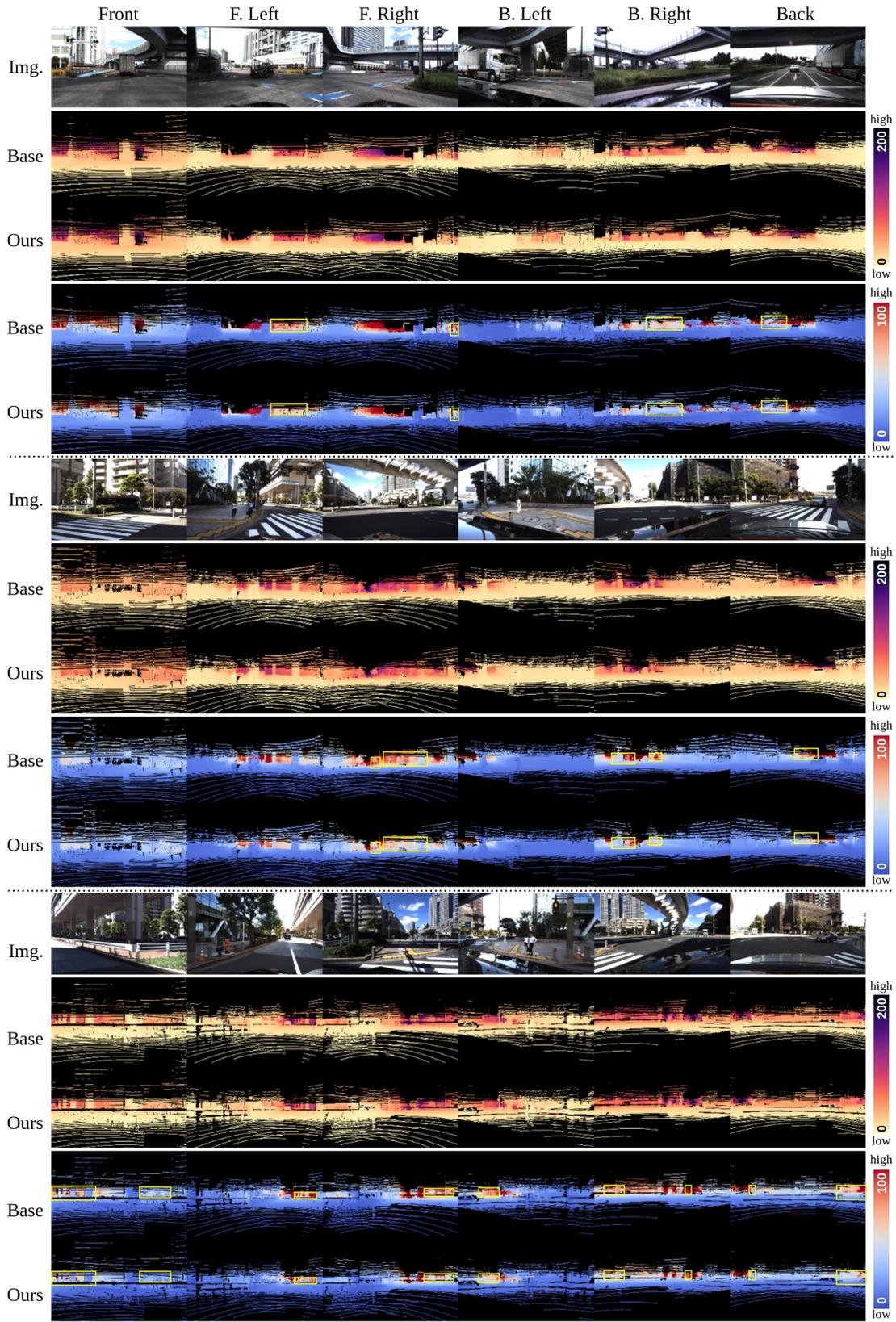}
    \caption{Additional qualitative results of fine-tuned MonoViT (denoted as Base) and MonoViT + Ours (denoted as Ours) on \textbf{DDAD} dataset. The second and third rows of each sample show the depth prediction, and the last two rows present the error map.}
    \label{fig:qualitative_ddad2}
\end{figure*}

\begin{figure*}[!t]
    \centering
    \includegraphics[width=0.90\linewidth]{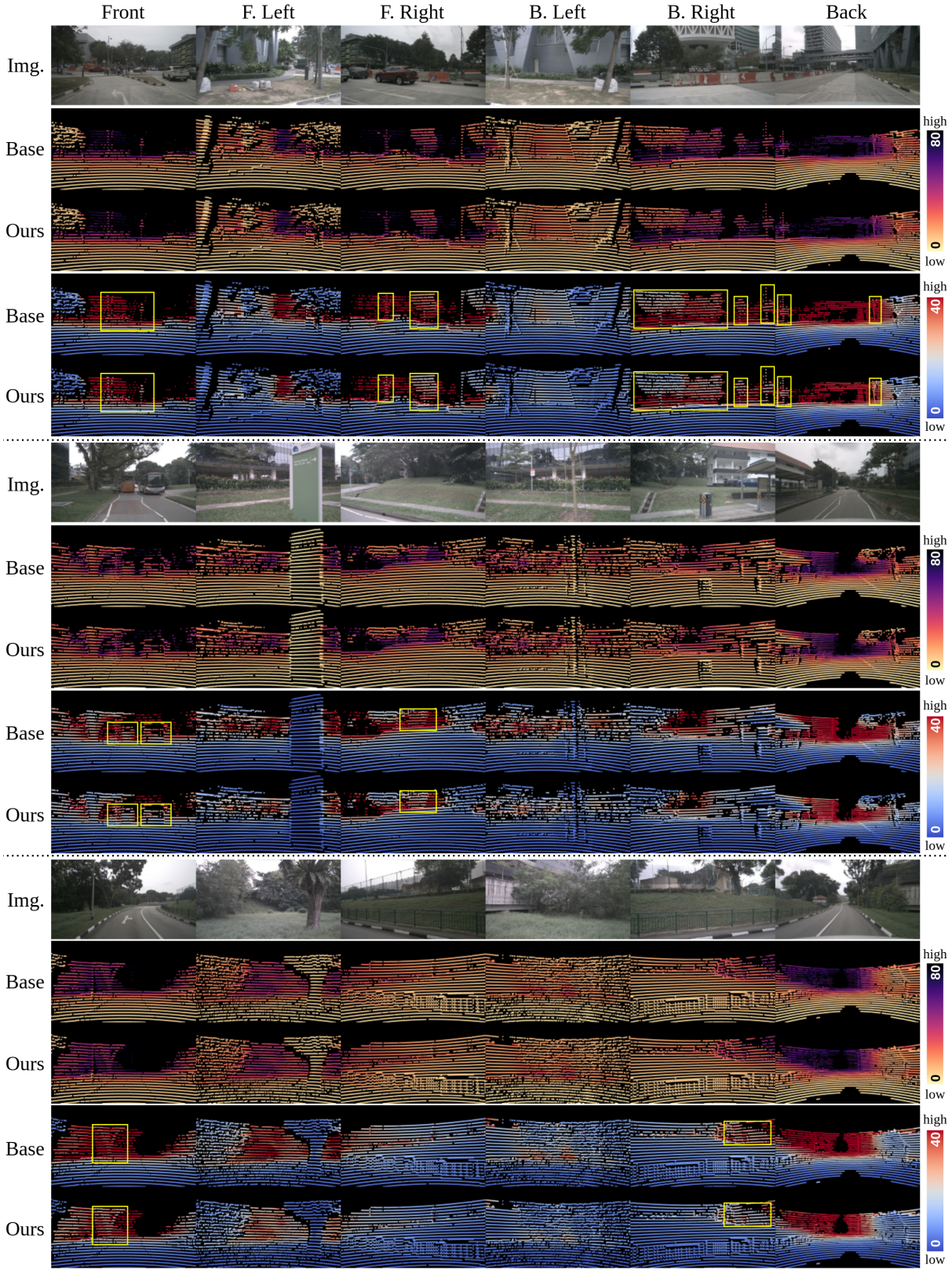}
    \caption{Additional qualitative results of fine-tuned MonoViT (denoted as Base) and MonoViT + Ours (denoted as Ours) on \textbf{nuScenes} dataset. The second and third rows of each sample show the depth prediction, and the last two rows present the error map.}
    \label{fig:qualitative_nuscenes2}
\end{figure*}

We provide additional qualitative results for DDAD and nuScenes, complementing the evaluation in \figref{fig:qualitative_ddad} of the main manuscript, using MonoViT and MonoViT + Ours.
As observed in \figref{fig:qualitative_ddad2}-\ref{fig:qualitative_nuscenes2}, our method consistently refines depth estimations, particularly in long-range regions, where standard monocular models tend to struggle.
Additionally, our approach effectively captures fine structural details, reducing errors in depth discontinuities and improving the overall visual quality of the estimated depth maps.

\section{Limitation \& Future work}
While our method demonstrates strong performance, certain limitations remain that warrant further investigation.
In this section, we discuss key challenges and potential directions to enhance its applicability and robustness.

\noindent\textbf{(1) Dependency on Teacher Model Binning:}
Our method relies on the binning strategy of the teacher model, limiting its applicability when binning is absent. Integrating a binning module into the foundation model could address this, but it adds computational overhead.

\noindent\textbf{(2) Sensitivity to Teacher Model Quality:}
The student model inherits biases and errors from the teacher.
Since our experiments focus solely on DepthAnything~\cite{yang2024depth}, evaluating other teacher models (e.g., Marigold~\cite{ke2024repurposing}, UniDepth~\cite{piccinelli2024unidepth}, and DepthPro~\cite{bochkovskii2024depth}) would help assess the generalizability of our approach.

% \noindent\textbf{(3) Justification for VRKD vs. Geometric Constraints:}
% Our View-relational Knowledge Distillation (VRKD) improves cross-view consistency, but its advantages over traditional geometric constraints (e.g., epipolar constraints) require further investigation.

\noindent\textbf{(3) Sparse Ground Truth in Supervised FSMDE:}
Projected LiDAR points lack information in certain areas (e.g., sky, and reflective surfaces), potentially leading to inaccuracies.
Future work could explore photometric consistency or depth completion techniques to mitigate this issue.

\clearpage

\twocolumn


%%%%%%%%%%%%%%%%%%%%%%%%%%%%%%%%%%%%%%%%%%%%%%%%%%%%%%%%%%%%%%%%%%%%%%%%%%%%%%%%

%%%%%%%%%%%%%%%%%%%%%%%%%%%%%%%%%%%%%%%%%%%%%%%%%%%%%%%%%%%%%%%%%%%%%%%%%%%%%%%%



% \section*{ACKNOWLEDGMENT}

% The preferred spelling of the word ÒacknowledgmentÓ in America is without an ÒeÓ after the ÒgÓ. Avoid the stilted expression, ÒOne of us (R. B. G.) thanks . . .Ó  Instead, try ÒR. B. G. thanksÓ. Put sponsor acknowledgments in the unnumbered footnote on the first page.

%%%%%%%%%%%%%%%%%%%%%%%%%%%%%%%%%%%%%%%%%%%%%%%%%%%%%%%%%%%%%%%%%%%%%%%%%%%%%%%%

% References are important to the reader; therefore, each citation must be complete and correct. If at all possible, references should be commonly available publications.
% \bibliographystyle{IEEEtran}
% \bibliography{IEEEexample}

\end{document}